\documentclass{article} 

\let\originaldelta\delta

\usepackage[final]{colm2025_conference}

\usepackage{times}
\usepackage{latexsym}
\usepackage{amsmath}
\usepackage{amssymb}
\usepackage{subcaption}
\usepackage{placeins}
\usepackage{subcaption}  

\usepackage{xcolor}
\usepackage{dsfont}
\usepackage{hyperref}
\usepackage{cleveref}
\usepackage{titlesec}
\usepackage{float}
\usepackage{tikz-dependency}
\usepackage{graphicx}
\usepackage{microtype}
\usepackage{url}
\usepackage{booktabs}
\usepackage{lineno}
\usepackage{wrapfig}
\usepackage{tcolorbox}
\usepackage{colortbl}      
\usepackage{authblk}           
\usepackage[symbol]{footmisc}   

\renewcommand{\thesubfigure}{%
  \ifcase\value{subfigure}%
  \or , top left%
  \or , top middle%
  \or , top right%
  \or , bottom left%
  \or , bottom middle%
  \or , bottom right%
  \else \arabic{subfigure}%
  \fi
}

\definecolor{darkblue}{rgb}{0, 0, 0.5}
\hypersetup{colorlinks=true, citecolor=darkblue, linkcolor=darkblue, urlcolor=darkblue}

\title{\raggedright
       Probing Syntax in Large Language Models:\\ Successes and Remaining Challenges}

\newcommand{\corrnote}[1]{\thanks{Correspondance to: \texttt{#1}}\enspace}

\author[1]{Pablo J.~Diego-Sim\'on\corrnote{pablo.diego-simon@psl.eu}}
\author[1,2]{Emmanuel Chemla}
\author[3]{Jean-R\'emi King}
\author[1]{Yair Lakretz}

\affil[1]{LSCP, ENS, PSL, EHESS, CNRS, Paris, France}
\affil[2]{Earth Species Project, Berkeley, California, USA}
\affil[3]{Meta AI, Paris, France}

\begin{document}

\ifcolmsubmission
\linenumbers
\fi

{\centering
\maketitle}

\begin{abstract}
The syntactic structures of sentences can be readily read-out from the activations of large language models (LLMs).
However, the ``structural probes'' that have been developed to reveal this phenomenon are typically evaluated on an indiscriminate set of sentences. Consequently, it remains unclear whether structural and/or statistical factors systematically affect these syntactic representations.
To address this issue, we conduct an in-depth analysis of structural probes on three controlled benchmarks. 
Our results are fourfold. 
First, structural probes are biased by a superficial property: the closer two words are in a sentence, the more likely structural probes will consider them as syntactically linked.
Second, structural probes are challenged by linguistic properties: they poorly represent deep syntactic structures, and get interfered by interacting nouns or ungrammatical verb forms. 
Third, structural probes do not appear to be affected by the LLMs' predictability of individual words. 
Fourth, despite these challenges, structural probes still reveal syntactic links far more accurately than the linear baseline or the LLMs' raw activation spaces.
Taken together, this work sheds light on both the challenges and the successes of current structural probes and provides a benchmark made of controlled stimuli to better evaluate their performance.

\end{abstract}

\section{Introduction}

\paragraph{The autonomy of syntax.} Understanding how word sequences are combined to form the meaning of a sentence is a central challenge to linguistics \citep{Chomsky1957,Chomsky1995}. 
Behavioral \citep{Rayner1978, Frazier1987, Gleitman1990} and neuroimaging research \citep{Pallier2011,Dehaene2011,Devauchelle2009,Caucheteux2021,Friederici, Santi2007} show that this phenomenon requires the brain to build latent structures that link the words of sentences. 
Critically, such ``syntactic'' system is thought to be \emph{autonomous}: the infamous sentence ``colorless green ideas sleep furiously'' shows that words can be syntactically linked even if each word is highly unpredictable, and thus lead to a nonsensical sentence \cite{Croft1995,Chomsky1957}. 
Yet, the neural and computational implementations of such \emph{autonomous} syntactic structures remain largely unresolved \citep{Grodzinsky, kaan2002}.

\paragraph{Syntactic representations in LLMs.} Large Language Models (LLMs) now offer an unprecedented opportunity to revisit this historical question. Not only can these models generate coherent sentences, but their internal activations explicitly represent the syntactic structures long theorized by linguists \citep{belinkov2017, Tenney2018, Peters2018, hale2024,Hewitt2019,DiegoSimon2024}. 
Following earlier work on linear probing \citep{Alain2017, conneau2018}, the `Structural Probe' \citep{Hewitt2019} recently showed that LLMs spontaneously learn to build a subspace of activations, where the distance between two contextualized word embedding corresponds to the distance that separates two words in the syntactic tree. \citet{DiegoSimon2024} further showed that the angle between these two words can represent the type of syntactic relations (e.g. `subject of', `object of'). 

\paragraph{Challenge. } However, the performance of such Syntactic Probes has \emph{mostly} been assessed with aggregated scores over large naturalistic corpora \citep{Guarasci2021}. 
Consequently,  
it is unclear whether the syntactic representations revealed by these LLM probes identify abstract representations of syntax \citep{Croft1995, Chomsky1957}, or whether they rather rely on other, heuristic or statistical factors.

\paragraph{Approach.} To address this question, we evaluate the performance of two state-of-the-art syntactic probes \citep{Hewitt2019,DiegoSimon2024} under a variety of linguistic conditions. To this end, we leverage both naturalistic datasets and controlled stimuli designed to manipulate specific linguistic (linear distance, depth, number interference) and statistical (word predictability) features. 
With these tests, we aim to (1)~characterize the strengths and limitations of structural probes
(2)~compare them with human behavior, and ultimately 
(3)~provide a benchmark to better evaluate structural probes.

\begin{figure*}[!t]
  \centering
  \includegraphics[width=0.99\textwidth]{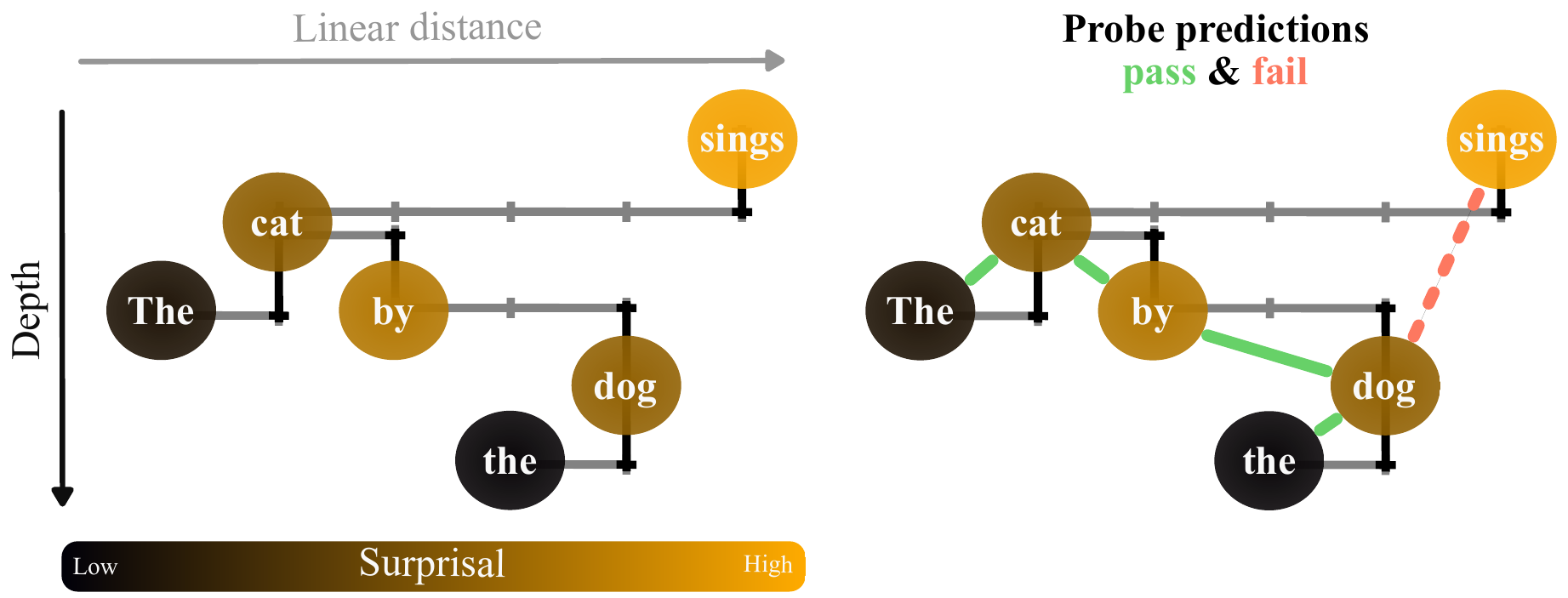}

    \caption{\textbf{Dependencies form a tree structure linking words in sentences, which the Structural Probe aims to predict.} 
    Each dependency connects two words that are separated by a certain \textit{linear distance}. Since the dependency tree is rooted, one word in the dependency serves as the head and the other as the child, introducing the notion of \textit{syntactic depth}. 
    Beyond these structural features, the LLM assigns each word a \textit{surprisal} value based on its downstream prediction. 
    We investigate how linear distance, depth, and surprisal influence probe predictions.}

  \label{fig:conceptual}
\end{figure*}

\section{Related work}

Understanding syntax with linear structural probes has been an extensive area of research initiated with the seminal work of \citet{Hewitt2019}. These probes have been extended using spectral \citep{MullerEberstein2022}, polar \citep{DiegoSimon2024}, and orthogonal \citep{Limisiewicz2021} transformations providing insights about the underlying vectorial spaces hosting syntactic processing \citep{Coenen2019}. 

This suite of probes rests on the premise that linguistic structure is linearly readable from LLM representations. Recent work suggests that linear representations also encode belief states, both across pre-training \citep{Shai2024} and during in-context learning \citep{Park2025}.

In parallel, efforts have been made to develop non-linear probes, which relax the linear constraint, offering a more flexible view of syntactic encoding \citep{Eisape2022, White2021, Chen2021}.

Probing methods, however, are subject to limitations and potential artifacts. Some have shown evidence that probe performance can be misleading, not generalizing to Jabberwocky sentences \citep{HallMaudslay2021} or having marginal effects on information-theoretical scores \citep{Pimentel2020}. Revealing that, probes may exploit shallow heuristics, making controlled and linguistically motivated evaluations essential for uncovering deeper insights.

\section{Methods}

\subsection{Data}

    \paragraph{Naturalistic sentences}{
    
    The Structural \citep{Hewitt2019} and Polar \citep{DiegoSimon2024} Probes were trained on the Universal Dependencies English Web Treebank (UD-EWT) corpus \footnote{\url{https://universaldependencies.org/treebanks/en_ewt/}} \citep{Silviera2014}. UD-EWT contains 16,622 sentences. The corpus consists of 254,820 words and 16,622 sentences, extracted from online journals, blogs and informative websites. Every sentence in the corpus has been manually annotated following the Universal Dependencies (UD) formalism \citep{Nivre2017}.
    
    Sentences that include email addresses or web URLs have been excluded from the dataset, as they are considered noisy and irrelevant for syntactic analysis. Also, sentences for which UD tokens and LLM tokens cannot be aligned are also excluded. We adhere to the default split provided by UD-EWT, resulting in 11,827 sentences for training, 1,851 for validation, and 1,869 for testing.}

    \paragraph{Controlled sentences} \label{par:controlled-sentences}
    We generated 3 controlled datasets one for each of the following syntactic structures:
    
      
      
    
    
    

    \noindent
    \begin{minipage}[t]{0.48\textwidth}
      \begin{tcolorbox}[colframe=black, arc=4mm, boxrule=0.5pt]
          \centering
          \textbf{Prepositional Phrase (PP)}
          
          \begin{dependency}
              \begin{deptext}
              The \& \textcolor{cyan}{mouse} \& by \& the \& \textcolor{olive}{cat} \& \textcolor{red}{moves}\\
              \end{deptext}
              \depedge[edge height=0.8ex]{6}{2}{subj}
          \end{dependency}
      \end{tcolorbox}
    \end{minipage}\hfill
    \begin{minipage}[t]{0.48\textwidth}
      \begin{tcolorbox}[colframe=black, arc=4mm, boxrule=0.5pt]
          \centering
          \textbf{Center Embedding (CE)}
          
          \begin{dependency}
              \begin{deptext}
              The \& \textcolor{cyan}{mouse} \& that \& the \& \textcolor{olive}{cat} \& chases \& \textcolor{red}{moves} \\
              \end{deptext}
              \depedge[edge height=0.8ex]{7}{2}{subj}
          \end{dependency}
      \end{tcolorbox}
    \end{minipage}
    
    \vspace{-0.5em} 
    
    \begin{center}
      \begin{tcolorbox}[colframe=black, arc=4mm, boxrule=0.5pt, width=0.55\textwidth]
        \centering
        \textbf{Right Branching (RB)}
        
        \begin{dependency}
            \begin{deptext}
            The \& \textcolor{cyan}{mouse} \& \textcolor{red}{believes} \& that \& the \& \textcolor{olive}{cat} \& moves\\
            \end{deptext}
            \depedge[edge height=0.8ex]{3}{2}{subj}
        \end{dependency}
      \end{tcolorbox}
    \end{center}

    In these syntactic structures, the \textcolor{olive}{attractor} interferes with the \textcolor{cyan}{subject} of the sentence. Furthermore, both PP and CE structures introduce long-range subject-verb  dependencies, in which the \textcolor{cyan}{subject} is separated from the \textcolor{red}{main verb}. Each dataset contains sentences with varying degrees of syntactic nesting, ranging from 1 to 3 levels. Adding a syntactic nesting modifies the syntactic depth of the sentence and introduces an \textcolor{olive}{attractor} noun.
    For the PP and the CE case, adding more nestings implies modifying the linear distance between the \textcolor{red}{head} and the \textcolor{cyan}{dependent} in the subject-verb dependency. For the 2 nesting case each structure becomes:
    
    \begin{itemize}
        \item \textit{The \textcolor{cyan}{mouse} by the \textcolor{olive}{cat} beside the \textcolor{olive}{fox} \textcolor{red}{moves}}
        \item \textit{The \textcolor{cyan}{mouse} that the \textcolor{olive}{cat} that the \textcolor{olive}{fox} protects chases \textcolor{red}{moves}}
        \item \textit{The \textcolor{olive}{cat} believes that the \textcolor{olive}{fox} expects that the \textcolor{cyan}{mouse} \textcolor{red}{moves}}
    \end{itemize}

    The controlled dataset is generated using a generative grammar tailored to each target syntactic structure: PP, CE, and RB. The number of nestings for each sentence is uniformly sampled from \(\{1, 2, 3\}\), introducing varying levels of syntactic complexity. Three nestings are already unusual in natural language and difficult for humans to understand. A shared vocabulary is used across all structures, consisting of five lexical categories: verb, noun, transitive verb, preposition, and determiner. Lexical items are sampled from these categories for each syntactic structure, with replacement allowed only for the determiner category. To maintain dataset integrity, we ensure that all generated sentences are unique.

    Finally, to dissociate the effects of additional attractors from those of increased linear distance between the \textcolor{cyan}{subject} and the \textcolor{red}{verb}, we augment the dataset simple sentences containing \textcolor{brown}{fillers}. In these constructions, the fillers are adverbs that modify the main verb. \textcolor{brown}{Fillers} extend the linear distance without introducing \textcolor{olive}{attractors}.

    \begin{itemize}
    \item Simple sentence [2 fillers]: \\
    \textit{The \textcolor{cyan}{mouse} \textcolor{brown}{quickly} and \textcolor{brown}{silently} \textcolor{red}{moves}.}
    \end{itemize}

   Overall, the controlled dataset consists of 80,000 sentences. To extend the analysis and generate ungrammatical sentences, the same sentences are used, but the main verb form is altered to mismatch the subject-verb agreement.


\subsection{Probes}
\paragraph{Structural Probe}
Let \( W = (w_1, w_2, \dots, w_t) \) be a sentence consisting of \( t \) words belonging to a dataset \( \mathcal{D} \). Let \( T_W = (V_W, E_W) \) be its corresponding syntactic dependency tree, where \( V_W \) is the set of words, and \( E_W \) is the set of edges connecting \( V_W \) in an acyclic manner. The adjacency matrix \( A \in \{0,1\}^{t \times t} \)  of \( T_W\) is defined as:

\begin{equation}
\mathbf{A}_{i,j} = 
\begin{cases} 
    1, & \text{if } \{w_i, w_j\} \in E_W, \\ 
    0, & \text{otherwise}.
\end{cases}
\end{equation}

For any two words \( w_i, w_j \in W \), the syntactic distance \( d_T(w_i, w_j) \) is defined as the number of edges in the unique path connecting \( w_i \) and \( w_j \) in \( T_W \). Formally, it can be written as:

\begin{equation} 
d_T(w_i, w_j) = \arg\min_k \left(\mathds{1}[(\mathbf{A}_{i,j})^k > 0] \right),
\end{equation} 

where \(\mathds{1}[\cdotp]\) represents the indicator function.

Thus we can construct the pairwise syntactic distance \( \mathbf{M} \in \mathbb{N}_0^{t \times t}\) between all words in a sentence.

\begin{equation} 
\mathbf{M}_{i, j} = d_T(w_i, w_j)
\end{equation}

Let \( \mathbf{h}^l_i \in \mathbb{R}^d \) denote the contextual embedding of the word \( w_i \) obtained from layer \( l \) of an LLM with a hidden dimensionality \( d \). In cases where a word is tokenized several, the word representation \( \mathbf{h}^l_i \) is computed as the average of its subtoken embeddings to ensure a one-to-one correspondence between words and embeddings. 

We follow \citet{Hewitt2019} and introduce a Structural Probe, defined by a linear transformation \( \mathbf{B}_S \in \mathbb{R}^{m \times d} \), where \( m \leq d \) represents the output dimensionality of the Structural Probe.

In the original embedding space \(\mathcal{H}\), the syntactic relations between two words \({w_i, w_j} \in W\) in a sentence W, are interpreted as the difference between their contextualized word embeddings.

\begin{equation}
    \mathbf{\originaldelta}^l(w_i, w_j) = \mathbf{h}^l_i - \mathbf{h}^l_j
\end{equation}

The Structural Probe maps word representations from \( \mathcal{H} \) to a new space \( \mathcal{S} \).

\begin{equation}
    \mathbf{s}^l(w_i, w_j) = \mathbf{B}_S \mathbf{\originaldelta}^l(w_i, w_j)
\label{Eq:g}
\end{equation}

The pairwise squared euclidean distance matrix \( \hat{\mathbf{M}} \in \mathbb{R}^{t \times t}\) between probed word representations for a given sentence \( W \) can be computed as follows:

\begin{equation}
    \hat{\mathbf{M}}_{ij} = ||\mathbf{s}^l(w_i, w_j)||^{2}_{2},
\label{Eq:D}
\end{equation}
where \(||\cdotp||\) represents L2 norm.

The goal of the Structural Probe is to find a subspace \( \mathcal{S} \) in which \( \hat{\mathbf{M}}_{ij} \) approximates \( \mathbf{M}_{ij} \). For that, the Structural Probe will be trained with the following objective.

\begin{equation}
    \begin{aligned}
        \mathcal{L}_S(\mathbf{B}_S) &= \frac{1}{|D|}\sum_{W \in D}\frac{1}{|W|^2} \sum_{(i,j)=1}^{|W|} \left| \mathbf{M}_{ij} - \hat{\mathbf{M}}_{ij} \right|, \\
        \mathbf{B}_S^* &= \arg \min_{\mathbf{B}_S} \mathcal{L}_S.
    \end{aligned}
\label{Eq:LS}
\end{equation}

\paragraph{Polar Probe}

Complete dependency trees are directed and labeled acyclic graphs rather than undirected and unlabeled. Thus, each edge (dependency) \(e \in E_W\) in the graph (dependency tree) is associated with two functions: \( U \) and \( C \), which determine its directionality and label (dependency type) respectively.\\

The Polar Probe \citep{DiegoSimon2024}, extends the Structural Probe and predicts a labeled and directed dependency tree, approximating \( U \) and \( C \) for each predicted syntactic edge. We extend the study to the Polar Probe since it augments \citet{Hewitt2019}'s framework with additional syntactic information while preserving a linearly readable, distance-based syntactic code. For more details about its implementation and training refer to \Cref{sec:Appendix}.

\subsection{Training}

The Structural and Polar probes were trained on activations from layer 16 of \texttt{Mistral-7B-v0.1} \citep{Jiang2023}, \texttt{Llama-2-7b-hf} \citep{Touvron2023}, and \texttt{BERT-large} \citep{Devlin2019}, as this layer has been shown to best encode dependency structures\footnote{\texttt{Mistral-7B-v0.1} and \texttt{Llama-2-7b-hf} are transformer-based LLMs with 32 layers and 7 billion parameters. \texttt{BERT-large} is a masked language model with 24 layers and 340 million parameters.} \citep{DiegoSimon2024}.


Training was conducted on the filtered UD-EWT corpus \citep{Silviera2014} using Stochastic Gradient Descent (SGD) with the Adam optimizer \citep{Kingma2015}. Probes were trained for 30 epochs with a batch size of 200 sentences and a learning rate of 0.005. For the Polar Probe, the hyperparameter \( \lambda \) was set to 10.0 as specified in \citep{DiegoSimon2024}.

\paragraph{Evaluation}
Following \citet{Hewitt2019}, for each sentence \(W\) in the test or controlled dataset, we use the trained probe \(\mathbf{B}_P^*\) or \(\mathbf{B}_S^*\) to compute its probed pairwise distance matrix \(\hat{\mathbf{M}}\). Then, we apply Kruskal’s algorithm \citet{Kruskal1956} using networkx \citep{networkx} to obtain the Minimum Spanning Tree (MST), \(\hat{E}_W\). Ignoring edge direction and labels, the probe performance on a given gold edge \(e \in E_W\) is measured by whether \(e\) appears in \(\hat{E}_W\):
\begin{equation}
\label{Eq:edge-accuracy}
\text{Accuracy}(e) \;=\;
\begin{cases}
1, & \text{if } e \in \hat{E}_W,\\
0, & \text{otherwise}.
\end{cases}
\end{equation}

\subsection{Baselines}


To gauge the effectiveness of the probes, we compare them against a strong baseline, modifying the definition of \(\hat{\mathbf{M}}\ \in \mathbb{R}^{t \times t}\). For more baselines, refer to \Cref{sec:Appendix}.

\begin{itemize}
        \item \textbf{Activation Space prediction}
    
        In this baseline, we use the raw LLM activations directly, without training or applying 
        any additional linear transformation.
        
        \[
            \hat{\mathbf{M}}_{ij} 
            \;=\; 
            ||\mathbf{\originaldelta}^l(w_i, w_j)||_2^2
        \]
    
        
        
    

\end{itemize}

\subsection{Surprisal calculation}

Surprisal, in causal language models, quantifies the information content of a word $w_i$ via its conditional probability given its previous context \citep{Shannon1948}. We compute word probabilities \(p(w_i)\) \citep{Pimentel2024}, from which surprisal \( I \) is derived.

\begin{equation}
I(w_i) = -\log(p(w_i))
\end{equation}

\subsection{Classifiers}

Ridge Regression and Random Forest models were trained to predict the probe’s output for ground truth positive instances \( (e \in E_W) \) using linguistic features such as head depth, head surprisal, child surprisal, and linear distance. These classifier models reveal which properties of an existing syntactic dependency are most relevant for probes to predict it correctly.

The Ridge classifer model was implemented via scikit-learn’s \citep{scikit-learn} \texttt{RidgeClassifier} with a regularization parameter 
\( \alpha = 100.0\). In parallel, the Random Forest classifier was configured using scikit-learn’s \texttt{RandomForestClassifier} with 100 estimators, a maximum tree depth of 10, and a minimum of 500 samples per split to mitigate overfitting.

Both models were evaluated using 5-fold cross-validation. For each fold, feature importance values were extracted from the Random Forest model and subsequently signed using the corresponding coefficients from the Ridge Regression model.

\section{Results}

\begin{figure*}[t]
    \centering
    \begin{subfigure}[b]{0.32\textwidth}
        \centering
        \includegraphics[width=\textwidth]{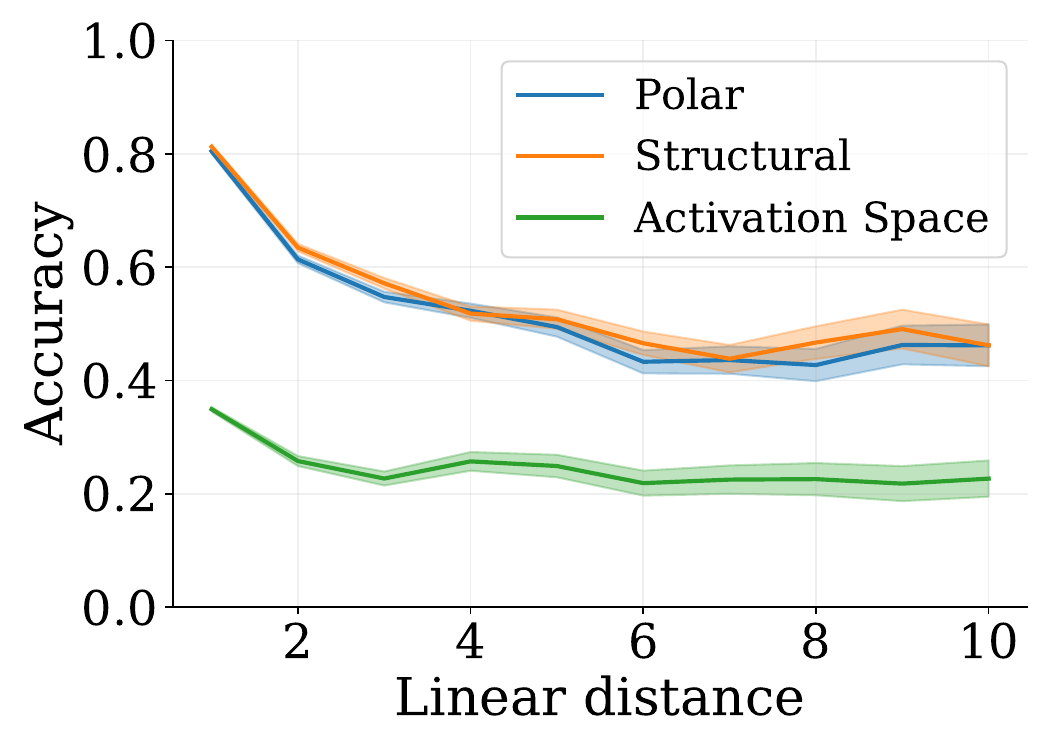}
        \phantomcaption
        \label{fig:lengths}
    \end{subfigure}
    \hfill
    \begin{subfigure}[b]{0.32\textwidth}
        \centering
        \includegraphics[width=\textwidth]{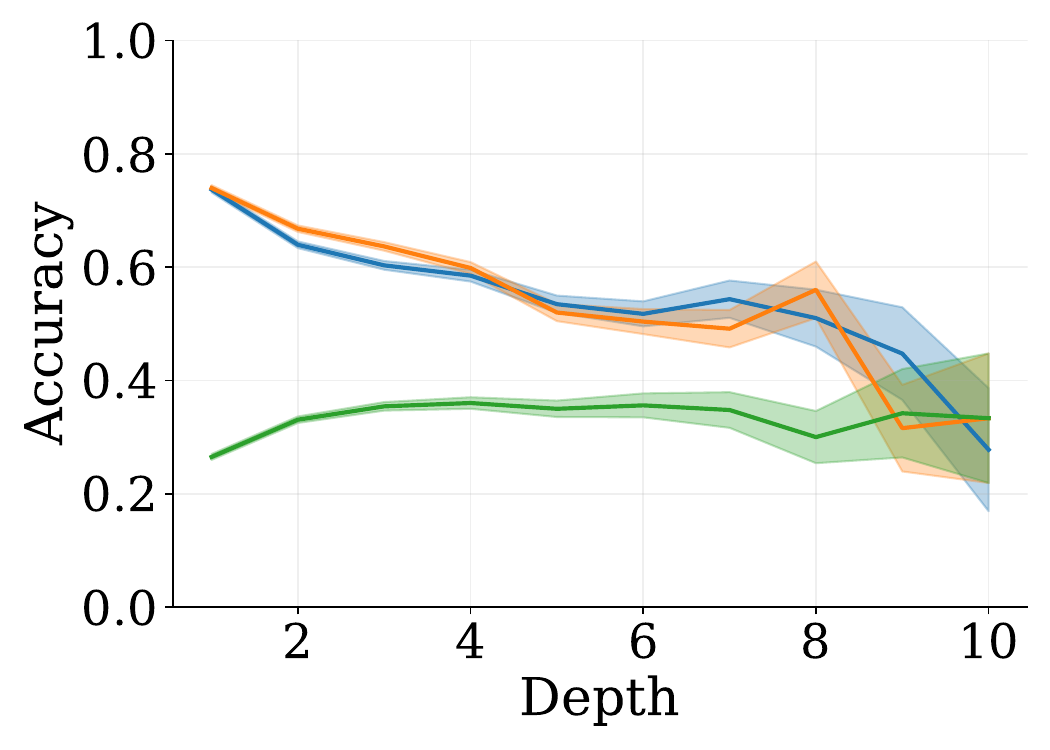}
        \phantomcaption
        \label{fig:depths}
    \end{subfigure}
    \hfill
    \begin{subfigure}[b]{0.32\textwidth}
        \centering
        \includegraphics[width=\textwidth]{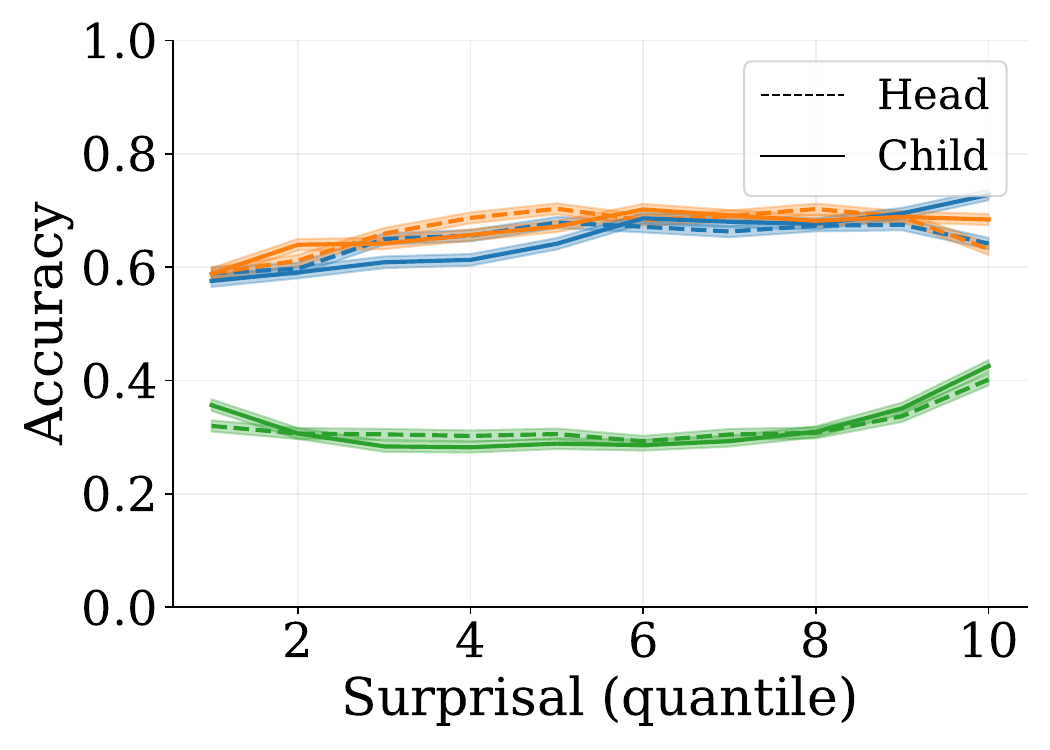}
        \phantomcaption
        \label{fig:surprisal}
    \end{subfigure}
    
    \vspace{-0.4cm} 
    
    \begin{subfigure}[b]{0.32\textwidth}
        \centering
        \includegraphics[width=\textwidth]{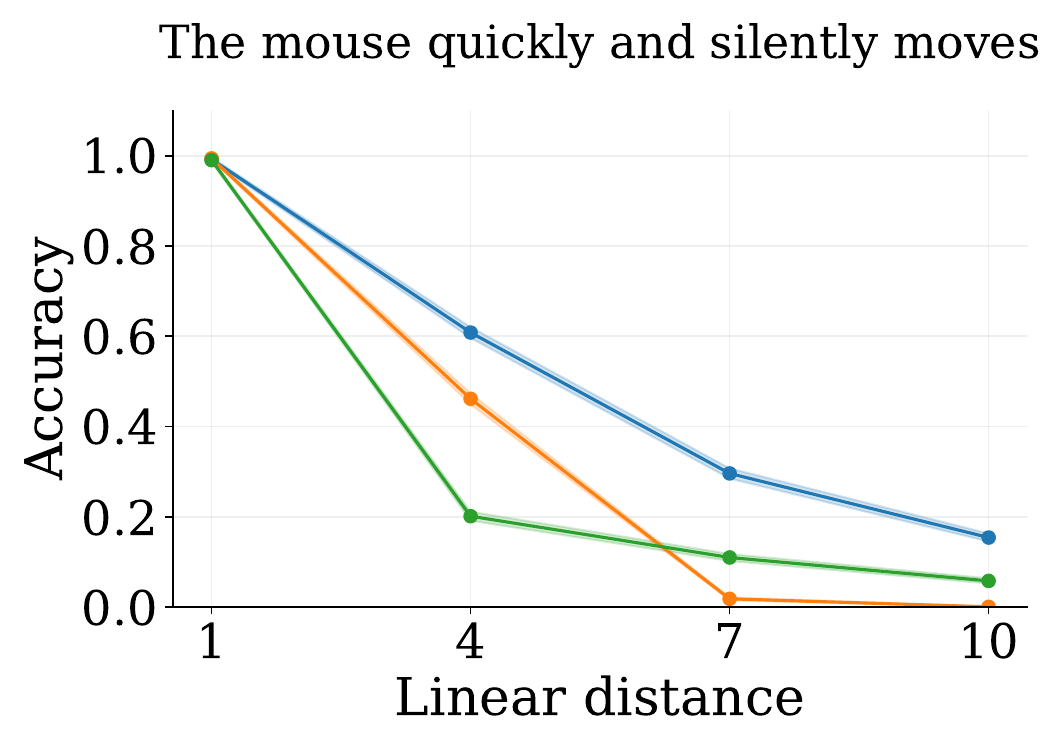}
        \phantomcaption
        \label{fig:lengths_controlled}
    \end{subfigure}
    \hfill
    \begin{subfigure}[b]{0.32\textwidth}
        \centering
        \includegraphics[width=\textwidth]{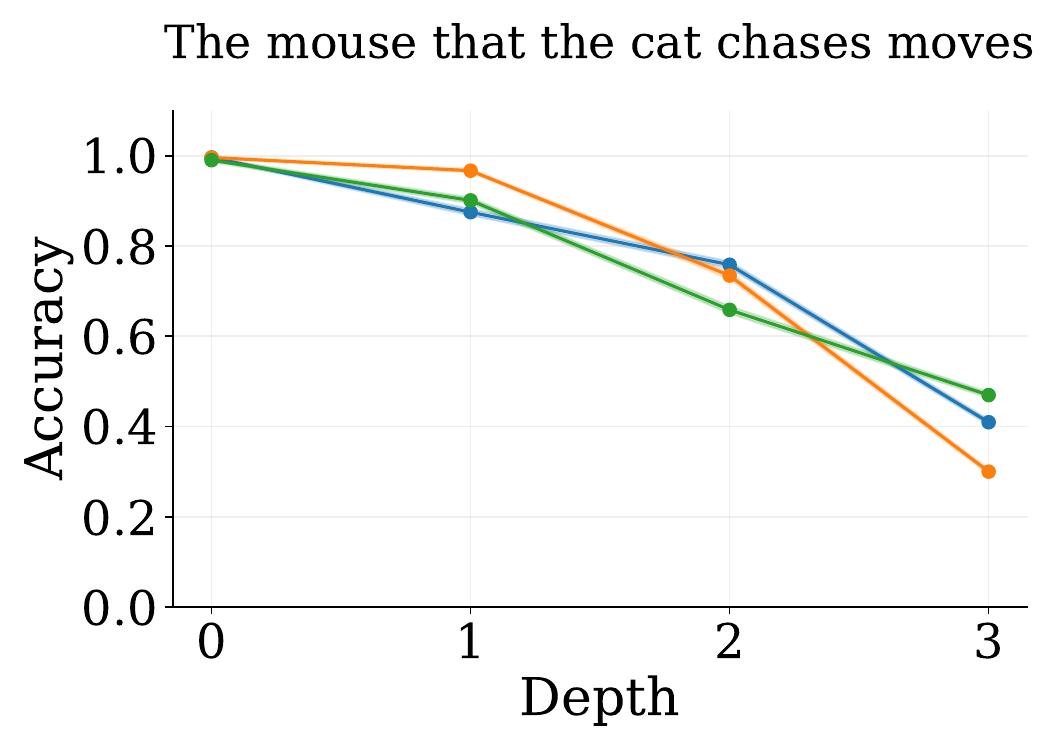}
        \phantomcaption
        \label{fig:depths_controlled}
    \end{subfigure}
    \hfill
    \begin{subfigure}[b]{0.32\textwidth}
        \centering
        \includegraphics[width=\textwidth]{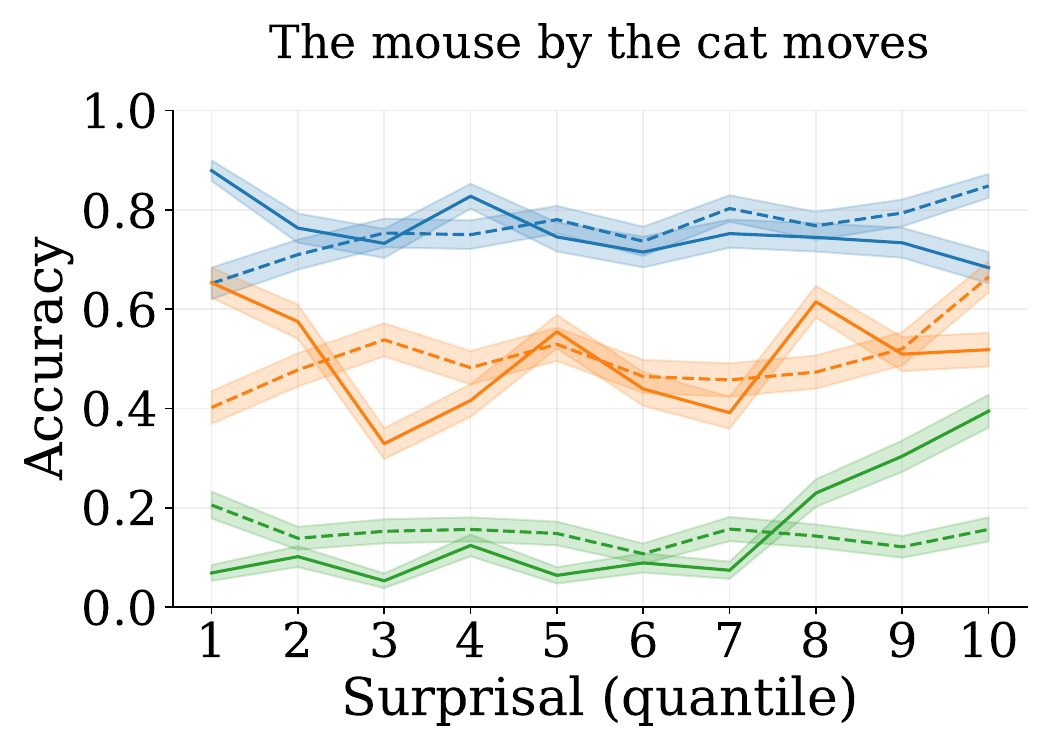}
        \phantomcaption
        \label{fig:surprisal_controlled}
    \end{subfigure}
    \vspace{-0.5cm}

    \caption{\textbf{The performance of trained linear probes is most affected by the linear distance and depth of syntactic dependencies in both naturalistic (Top) and controlled (Bottom) stimuli.} \textbf{(Top)}: Accuracy on syntactic dependencies in UD-EWT. \textbf{(Top Left)} By the linear distance between head and child. \textbf{(Top Middle)} By the depth of the head. \textbf{(Top Right)} By the surprisal quantile of the head and the child. \textbf{(Bottom):} Accuracy on the subject-verb dependency in controlled sentences. \textbf{(Bottom Left):} Simple sentences with a varying number of fillers. \textbf{(Bottom Middle)}: In CE sentences with varying number of nestings, modifying the syntactic depth of the innermost subject-verb dependency . \textbf{(Bottom Right):} In sentences with 1 PP by the surprisal quantile for head and child. Shaded areas indicate the standard error of the mean in all plots.}
    
    \label{fig:marginal_acc}
\end{figure*}

This section is structured as follows: First, we study which linguistic and statistical factors best predict probe performance (section \ref{ssec:probe_perf}), using both naturalistic (section \ref{sssec:naturalistic_data}) and manually-crafted controlled (section \ref{sssec:control_data}) datasets. Next, we investigate whether sentence parsing by structural probes exhibits processing phenomena similar to those observed in humans (section \ref{ssec:human_parse}).




\subsection{Probe performance is impaired by longer and deeper dependencies, but not by the surprisal of its elements.}\label{ssec:probe_perf}

\subsubsection{Naturalistic data}\label{sssec:naturalistic_data}

We first study the Structural and Polar Probes on a naturalistic dataset, the Universal Dependencies English Web Treebank (UD-EWT) dataset. For each analysis, we evaluate whether the distances between contextualized word embedding within the probe's subspace, accurately represents the syntactic distance in the annotated dependency tree.  
We evaluate the probe accuracy as a function of three linguistic features: 
(1) the linear distance that seperate two syntactically related words (i.e. the head and the dependent, \cref{fig:lengths}),
(2) the syntactic depth of the head (\cref{fig:depths}), 
and (3) the surprisal of the head (\cref{fig:surprisal}).

We find that both linear distance and syntactic depth negatively impact probe accuracy, indicating that long-range and deeply-nested dependencies still challenge current probes. Remarkably, for very deep structure (depth=10), the probe's accuracy is similar to what can be read out from the original space of activations of the LLM (\cref{fig:depths}).
By contrast, word surprisal shows little to no impact on probe accuracy. These results mean that the predictability of a word given its context does not affect the syntactic representation identified by the syntactic probes (\cref{fig:surprisal}).
Overall, these three results suggest that probe performance is more sensitive to structural properties of language than to statistical predictability. Suggesting a better performance generalization to nonsensical sentences rather than to sentences containing long-range or deep dependencies.


\begin{wrapfigure}{r}{0.55\textwidth}
  \centering
  \vspace{-5pt}
  \includegraphics[width=0.54\textwidth]{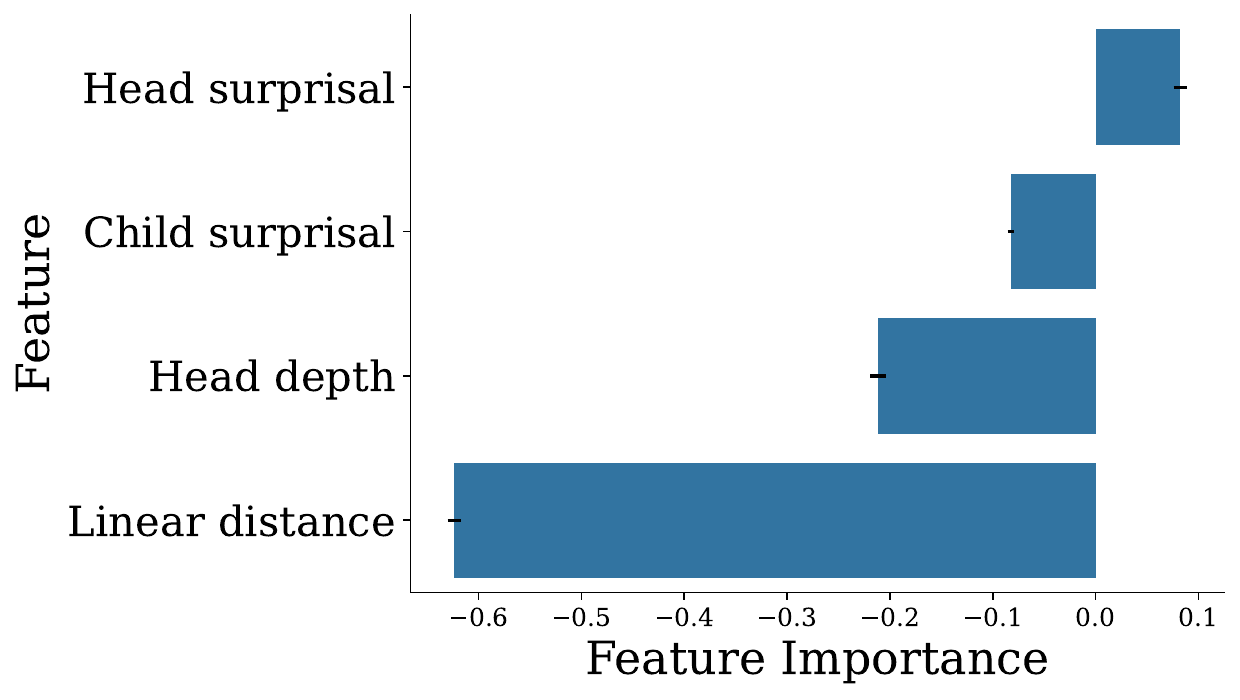}
  \caption{\textbf{Linear distance and depth of the syntactic dependency are the best predictors of probe performance} Signed mean feature importance in a Random Forest classifier trained to predict whether the Polar Probe detects existing dependencies in the UD-EWT dataset. The error bars represent the standard deviation of the feature importance values across 5 cross-validation folds.}
  \label{fig:coeff}
  \vspace{-20pt}
\end{wrapfigure}

 Linear distance, syntactic depth and word surprisals may not be fully independent from one another. To verify that the above results, we trained a Random Forest classifier to predict the probe decisions, depending on these three linguistic features (linear distance, syntactic depth and (head and child) surprisal).

\cref{fig:coeff} shows the signed feature importance for the different input variables.
Overall, the results confirm that probe accuracy are most affected by the linear distance and the syntactic depth. By contrast, surprisals lead to weaker and less consistent impact (\cref{fig:surprisal}). These results reinforce our earlier findings, suggesting that probe performance is most impacted by both syntactic depth and linear distance.

\subsubsection{Controlled sentences}\label{sssec:control_data}
The above analyses are based on syntactically annotated naturalistic sentences. To independently assess the impact linear distance, syntactic depth and word surprisal on the probes' accuracy, we extend our analyses to a set of controlled sentences, designed such that their syntactic dependencies systematically vary in linear distance and syntactic depth (see \cref{par:controlled-sentences}).
In all cases, probe accuracy is evaluated on the subject-verb dependency to ensure consistency across conditions. To manipulate linear distance (\cref{fig:lengths_controlled}), we used main phrase sentences with a varying number of adverbs as fillers (e.g., "The cat quickly and silently walks"). For syntactic depth (\cref{fig:depths_controlled}), we evaluate the innermost subject-verb dependency in CE sentences with different levels of nesting (e.g., "The cat that the fox chases moves"). Where the subject and verb lie at a linear distance of 1 for the different numbers of nesting. For surprisal (\cref{fig:surprisal_controlled}), we analyzed sentences with a single PP to avoid fillers (favoring more natural constructions) while introducing a long-range agreement.\\
Once again, our findings align with those from the UD-EWT dataset: linear distance and syntactic depth strongly impact probe accuracy. For linear distance, \cref{fig:lengths_controlled} we find a linear distance of 3 words is enough to decrease probe accuracy by 0,4 points. A similar effect happens for syntactic depth \cref{fig:depths_controlled}. Notably, for CE sentences, the `Activation Space' baseline matches with the Structural and Polar probes. We suggest that this is due to the subject and the verb being located at adjacent locations. Finally, for surprisal (\cref{fig:surprisal_controlled}), we observe opposing trends, as reflected in \cref{fig:coeff}, where accuracy slightly increases with head surprisal, whereas child surprisal exhibits the opposite effect. When comparing the Structural and Polar probes, we find that the Polar Probe exhibits a slower decay in accuracy with increasing linear distance \cref{fig:lengths_controlled}. We hypothesize that its more constrained objective, including additional syntactic information, fosters learning more robust syntactic representations, that rely less on surface heuristics.\\

\begin{figure*}[t]
  \centering
  \begin{subfigure}[b]{0.48\textwidth}
    \includegraphics[width=\textwidth]{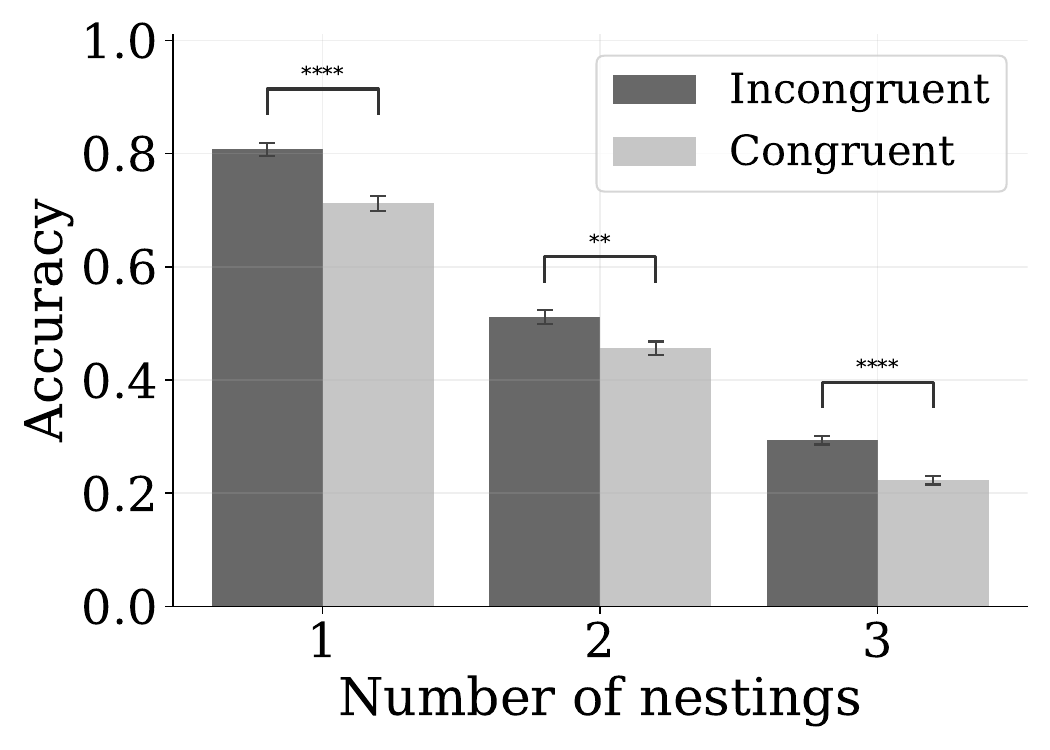}
    \phantomcaption
    \label{fig:cong_linear}
  \end{subfigure}
  \hfill
  \begin{subfigure}[b]{0.48\textwidth}
    \includegraphics[width=\textwidth]{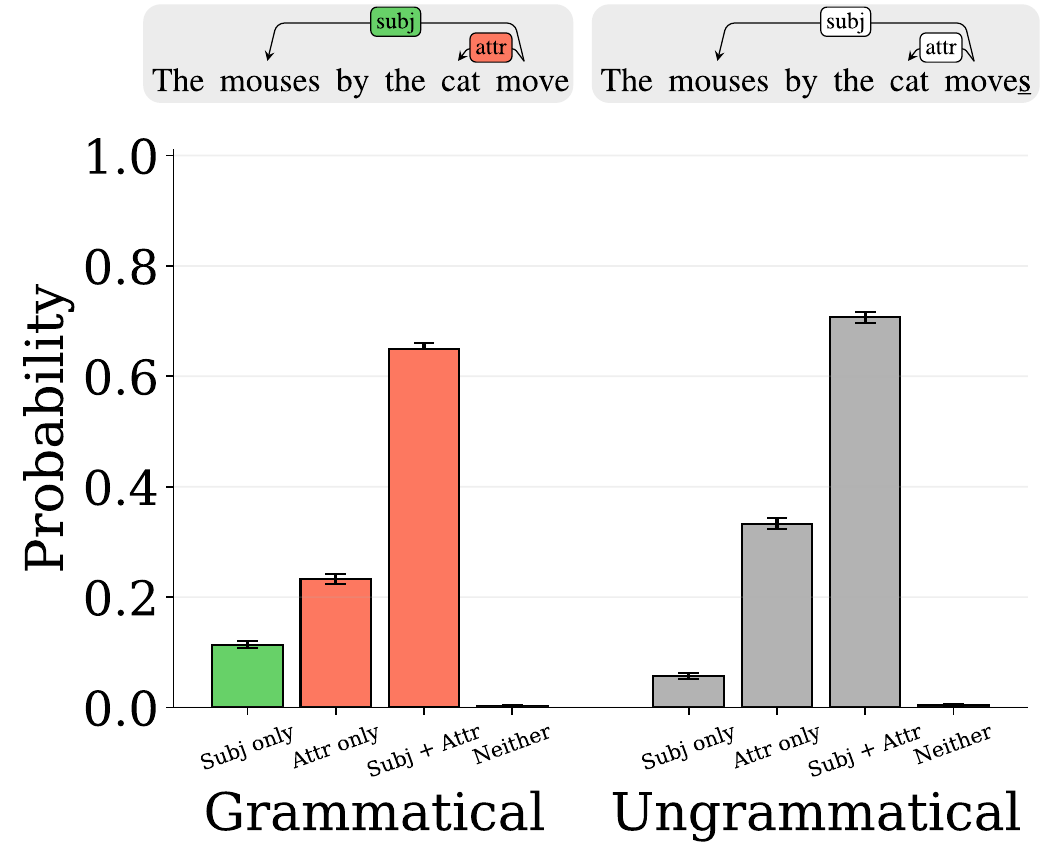}
    \phantomcaption
    \label{fig:flips}
  \end{subfigure}
  
  \vspace{-0.45cm} 
  
  \begin{subfigure}[b]{0.48\textwidth}
    \includegraphics[width=\textwidth]{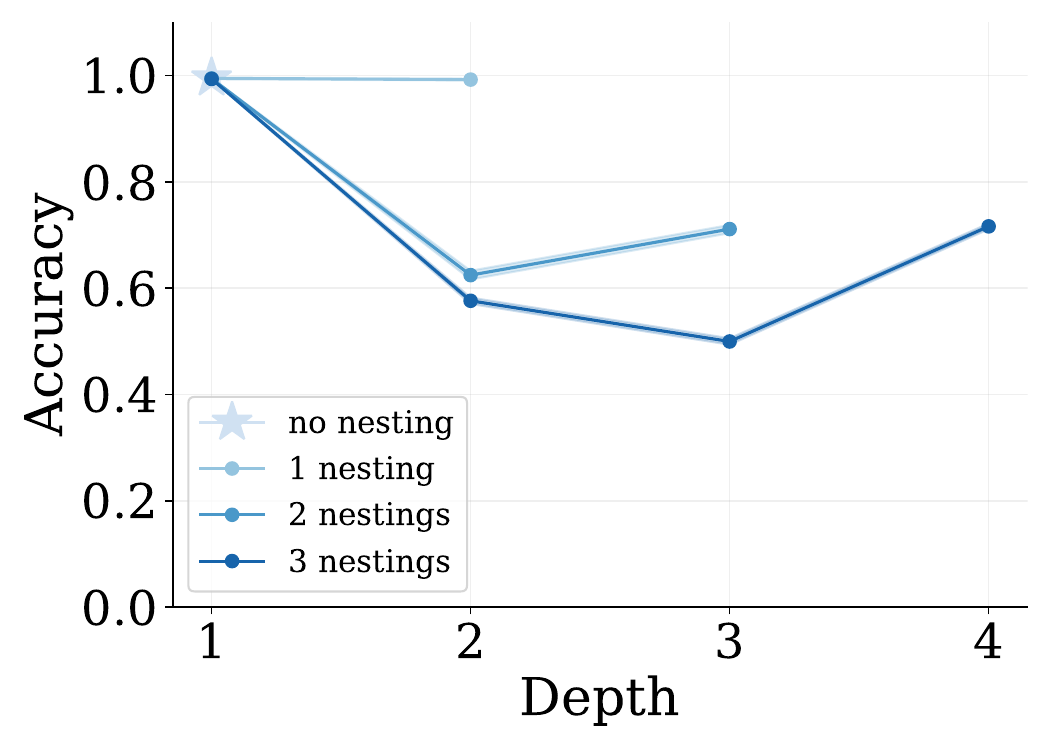}
    \phantomcaption
    \label{fig:controlled_depths_rb}
  \end{subfigure}
  \hfill
  \begin{subfigure}[b]{0.48\textwidth}
    \includegraphics[width=\textwidth]{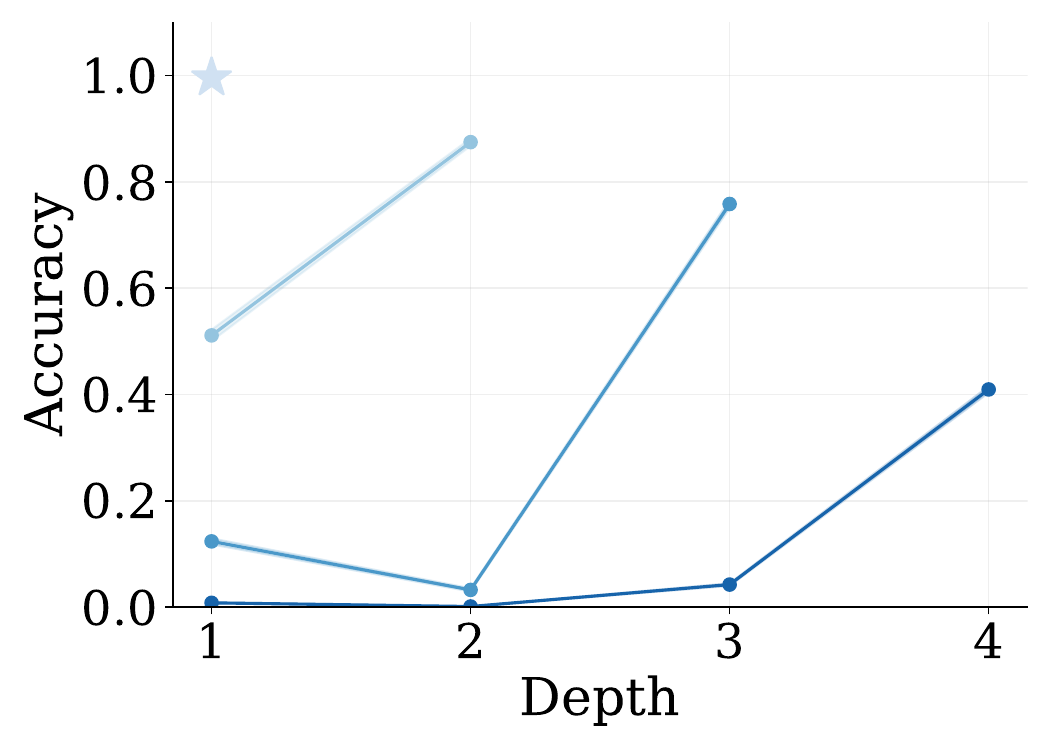}
    \phantomcaption
    \label{fig:controlled_depths_ce}
  \end{subfigure}
  
  \vspace{-0.5cm}
  \caption{\textbf{Probes share some similarities with human syntactic processing but fail on simple PP sentences.} \textbf{(Top left):} Accuracy of the Polar Probe on the subject-verb dependency in grammatical PP sentences with varying nesting levels, comparing cases where the final attractor is congruent or incongruent with the subject. Congruent cases are harder for the probe. \textbf{(Top right):} Probability of different Polar Probe predictions in grammatical and ungrammatical with one PP. Predictions are grouped by which word; the attractor, the subject, both, or neither—binds to the verb. Because the attractor adjacent to the verb, the probes are misled and produce mostly incorrect parses (red), outnumbering correct ones (green). Ungrammatical cases increase the attraction effect. \textbf{(Bottom left):} Accuracy profile for Right Branching (RB) where the linear distance is constant across nestings, modifying syntactic depth. Probes show primacy and recency effects. \textbf{(Bottom Right):} Accuracy profile for Center Embedding (CE) sentences, inner dependencies, despite being deeper, are also closer together resulting in higher probe accuracies. Error bars represent the standard error in all plots.}
  \label{fig:2x2_figure}
\end{figure*}

\subsection{Do probes and humans show similar parsing effects?}\label{ssec:human_parse}

Psycholinguistic research  \citep{Wagers2009, Bock1991, Francis1986, Kinball, Haskell2005} has shown that interfering nouns tend to bias individuals toward misinterpreting and misproducing long-range subject-verb agreements. These effects are especially pronounced in ungrammatical sentences, where attraction errors are more likely to occur. Moreover, increasing the linear distance between the subject and verb, as seen in prepositional phrase and object relative clauses, further generates confusion in syntactic agreement \citep{Bock1992}.

\subsubsection{Congruent nouns impact negatively on probe performance}

More recent work \citep{Lakretz2023, Lampinen2024, marvin-linzen-2018} has shown that congruent attractor nouns reinforce the confidence of LLMs when predicting verb forms during the next-token prediction task. (refer to \Cref{sec:Appendix} for more details regarding congruence and model prediction). We hypothesize that, in contrast, incongruent attractor nouns may cause confusion in the syntactic parsing task, which is not generative.

To test this hypothesis, measure sensitivity of probes to the congruency between the subject and the last attractor in PP sentences. Therefore, in congruent cases, the last attractor functions as an interfering noun positioned adjacent to the verb.

As depicted in \Cref{fig:cong_linear}, the Polar Probe exhibits greater difficulty with congruent cases, suggesting a tendency to associate the verb with the attractor when surface forms are compatible. Conversely, in incongruent cases, the absence of interference aids in disambiguating the correct subject-verb dependency, leading to improved probe performance. Notably, as indicated by the star notation in \Cref{fig:cong_linear}, the difference between the two groups is highly significant across the different levels of nesting. Lastly, \Cref{fig:cong_linear} illustrates the impact of attaching additional PPs on subject-verb accuracy; as expected, accuracy decreases with the number of PPs due to both the greater linear distance and the presence of more attractors.

\subsubsection{Ungrammatical verb forms impact negatively on probe performance}

Analogous to the interfering effects observed in grammatical sentences, previous works \citep{marvin-linzen-2018, Hu2024, ryu2021} have shown that ungrammaticality also fosters attraction effects in LLMs when evaluated in a generative setting. Mirroring findings from psycholinguistic studies. As illustrated in \Cref{fig:flips,fig:flips_llama}, we find that in PP sentences with varying levels of nesting, the Polar Probe exhibits a consistently higher error rate when encountering ungrammatical verb forms. 

\subsubsection{One PP suffices to elicit attraction effects}
\label{sssec:PP_fail}

Remarkably, we note that even in simple sentences with a single PP, the Polar Probe tends to bind the verb to both the subject and the attractor, resulting in a wrong parse (\Cref{fig:flips,fig:flips_llama}).

Such binding strategy comes at the expense of accurately capturing the \emph{case} relation within the PP. Surprisingly, the \emph{case} relation is challenging for the probe to parse as revealed by further dissection the UD-EWT dataset into dependency types. Refer to \Cref{sec:Appendix} for more details. Overall we find such result as an indicator of the extent to which probes are sensitive to linear distance. Such sensitiveness comes with prediction errors and points a solid separation between humans and probes.

As shown in \Cref{fig:flips_bert-large} such failure is somewhat mitigated by the use of a masked language model, where attention heads operate bidirectionally.

\subsubsection{Accuracy profiles for Right Branching and Center Embedding}

Controlled sentences with Right Branching (RB) and Center Embedding (CE) structures each include one subject-verb dependency per level of nesting. In RB sentences, syntactic depth increases while the linear distance between the subject and the verb remains constant. In contrast, in CE sentences, the subject-verb linear distance decreases as syntactic depth increases. These structural differences lead to distinct accuracy profiles with respect to the number nestings and syntactic depth.

As shown in \cref{fig:controlled_depths_rb} and consistent with the results presented in \cref{fig:marginal_acc}, accuracy in RB sentences declines with increasing syntactic depth. Notably, a U-shaped accuracy profile emerges, suggestive of primacy and recency effects documented in human cognition \citep{Murdock1962, Atkinson1968, Bower}, where items at the beginning and end of a sequence tend to be processed more accurately than those in the middle.

In contrast, accuracy in CE sentences improves with deeper nesting. This finding underscores the dominant influence of linear distance over syntactic depth as shown in \cref{fig:coeff}.

\section{Discussion, Limitations and Future work}

Our analyses reveal that accuracy of syntactic probes is primarily sensitive to structural rather than to statistical properties. In fact, in both naturalistic and controlled datasets, we find that the model's surprisal to the words in the dependency barely has an impact on probe accuracy. In contrast, probe accuracy decreases as the linear distance and syntactic depth of the dependency increases. 

Notably, in relatively simple sentences with one Prepositional Phrase (eg: The keys to the cabinet are big.), the probe tends to bind both the subject and the attractor to the verb, yielding a wrong parse (\Cref{sssec:PP_fail}). 

Such result puts forward a key challenge for linear probes, their sensitiveness to linear distance and syntactic depth makes them prompt to errors. In the PP case, the probes' sensitiveness to linear distance becomes especially apparent— the attractor being adjacent to the main verb is enough to confuse probes to yield a wrong syntactic parse.

To address these challenges, and better evaluate probes, we introduce a set of controlled sentences to serve as a benchmark. These carefully designed stimuli enable a linguistically motivated and systematic evaluation of probe performance.

Despite these challenges, probes also exhibit notable successes. They capture syntactic structure more accurately than the model’s raw activation space, and perform far beyond a linear distance baseline. Moreover, we find some similarities between probes and human behavior. Specifically, these similarities manifest as increased syntactic errors under both noun interference and presence ungrammatical verb forms \citep{Wagers2009, Bock1991}. However, this resemblance remains superficial due to the current failure cases observed in linear structural probes. 

Several factors likely underlie the current challenges of structural probes. First, their training data is strongly skewed toward dependencies with a linear distance of one, introducing a bias into the learned linear probe during its training. Second, if LLMs encode syntax along a nonlinear manifold in hidden space, any strictly linear probe will inevitably distort that representation. Third, some errors may originate not from the probes themselves but from the LLMs, which may fail to encode syntactic structure with full fidelity. Lastly, prompting LLMs has been shown to affect their downstream behavior \citep{Lampinen2024}, contextualizing the sentences using a carefully designed prompt might result in linguistically richer representations, potentially enhancing the performance of syntactic probes.

We posit that exploring non-Euclidean probes that better align with the geometry of LLM representations is a promising direction for future work. Likewise, training syntactic probes on controlled stimuli to better capture long-range dependencies and deeper hierarchical structures remains an important area for continued research.


\section{Acknowledgments}
This project was provided with computer and storage resources by GENCI at IDRIS thanks to the grant 2023-AD011014766 on the supercomputer Jean Zay’s the V100 and A100 partition (PDS).\\
This project has received funding from the European Union’s Horizon 2020 research and innovation program under the Marie Sklodowska-Curie grant agreement No 945304 (PDS).\\
This project received funding from PSL University under the grant agreement ANR-10-IDEX-0001-02 (EC).\\
This project received funding from the Département d'Études Cognitives (DEC) at ENS under the grant agreement FrontCog, ANR-17-EURE-0017 (EC).\\
This project received funding from the French National Research Agency (ANR) under the grant agreement ComCogMean, Projet-ANR-23-CE28-0016 (EC).




\bibliography{colm2025_conference}
\bibliographystyle{colm2025_conference}

\newpage
\newpage
\clearpage

\appendix
\section{Appendix}\label{sec:Appendix}

\subsection{Polar Probe implementation}

Each edge (dependency) \(e \in E_W\) in the graph (dependency tree) is associated with two functions: \( U \) and \( C \), which determine its directionality and label (dependency type) respectively.

\begin{equation}
    \begin{aligned}
        U: E_W &\to \{-1, 1\}, \quad e \mapsto u(e), \\
        C: E_W &\to C, \quad e \mapsto c(e).
    \end{aligned}
\end{equation}

Similarily to the Structural Probe, the Polar Probe \citet{DiegoSimon2024} defined as \( \mathbf{B}_P \in \mathbb{R}^{m \times d} \) linearly transforms the LLM's embedding space \(\mathcal{H}\) into a subspace \(\mathcal{P}\) following ~\cref{Eq:g,Eq:D,Eq:LS}. 

\begin{equation}
    \begin{aligned}
        \mathbf{p}^l(w_i, w_j) &= \mathbf{B}_P \mathbf{\originaldelta}^l(w_i, w_j), \\
        \hat{\mathbf{M}}_{ij} &= ||\mathbf{p}^l(w_i, w_j)||^{2}_{2}
    \end{aligned}
\label{Eq:LP1}
\end{equation}

Additionally to the structural objective (~\cref{Eq:LS}), the Polar Probe introduces an angular objective \(\mathcal{L}_A\) so that \( \mathbf{p}^l(w_i, w_j)\) where \(\{w_i, w_j\} \in E_W\) encodes information about \( U \) and \( C \). For that, a set of edges \( \Omega_e\)  are extracted across sentences from dataset \( \mathcal{D}\). Then, the following objective is minimized:

\begin{equation}
\begin{split}
\mathcal{L}_A(\mathbf{B}_P) = \frac{1}{|\Omega_e|} \sum_{e, e' \in \Omega_e} 
\Big( \measuredangle(\mathbf{p}^l_e \, u(e), \mathbf{p}^l_{e'} \, u(e')) \\
- \mathds{1}[c(e) = c(e')] \Big)^2.
\end{split}
\label{Eq:LP2}
\end{equation}

where \( \measuredangle(\cdot, \cdot) \) denotes the cosine similarity, and we write \( \mathbf{p}^l_e \) as a shorthand for \( \mathbf{p}^l(w_i, w_j) \) for an edge \( e = \{w_i, w_j\} \in E_W \subset \Omega_e \), by slight abuse of notation.

Therefore, the Polar Probe is the result of the following objective function:

\begin{equation}
\mathbf{B}_P^* = \arg \min_{\mathbf{B}_P} \mathcal{L}_S(\mathbf{B}_P) + \lambda\mathcal{L}_A(\mathbf{B}_P)
\end{equation}

\subsection{Additional baselines:}

\begin{itemize}
    
        
    
        \item \textbf{Linearly informed random prediction}
        \label{control:lirt}
        
        In this baseline, for two words at positions \(i\) and \(j\) in a sentence, 
        we define their distance to be the absolute difference in their positions, 
        plus a small random perturbation \(\epsilon_{ij}\):
        \[
            \hat{\mathbf{M}}_{ij} 
            \;=\; 
            |i - j| \;+\; \epsilon_{ij}.
        \]
        Here, \(\epsilon_{ij}\) is drawn i.i.d.  \(\epsilon_{ij}\) is sampled i.i.d.\ for \(i < j\) from a noise distribution and we set  \(\epsilon_{ij} =\epsilon_{ji}\) and \(\epsilon_{ii} =0 \) to keep \( \hat{\mathbf{M}} \) symmetric and zero-diagonal.
        
        \item \textbf{Random prediction} 
    
        To establish a lower bound, we assign completely random distances:
        \[
            \hat{\mathbf{M}}_{ij} 
            \;=\;
            \epsilon_{ij},
        \]
\end{itemize}


\begin{figure}[H]
  \centering
  \includegraphics[width=0.5\columnwidth]{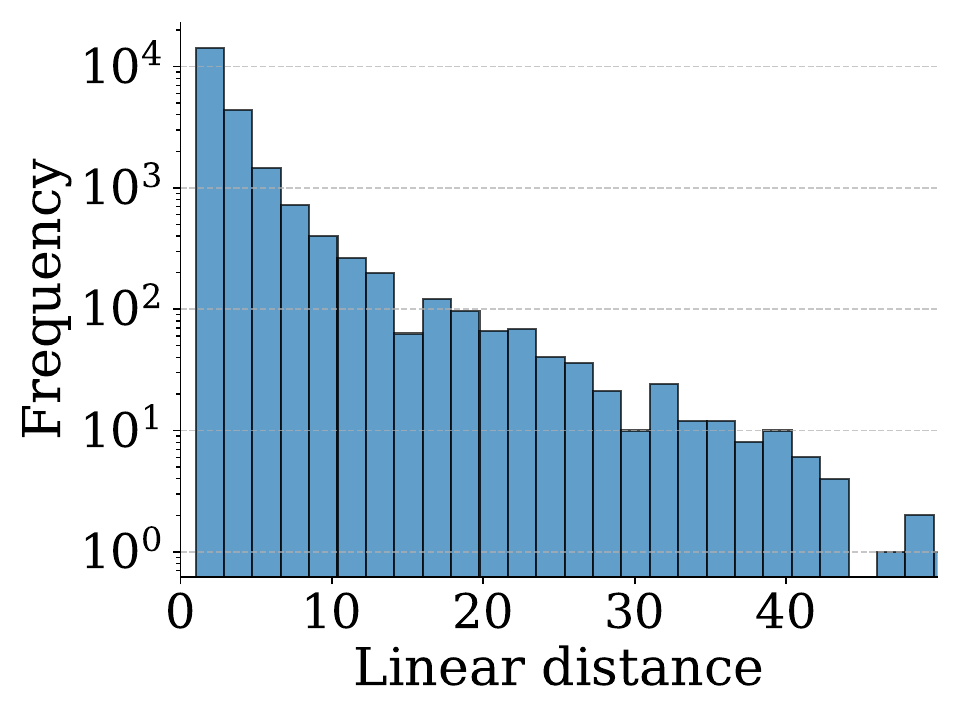}
  \caption{Histogram of linear distances for dependencies in the naturalistic sentences}
  \label{fig:hist}
\end{figure}

\begin{figure}[H]
  \centering
  \includegraphics[width=0.5\columnwidth]{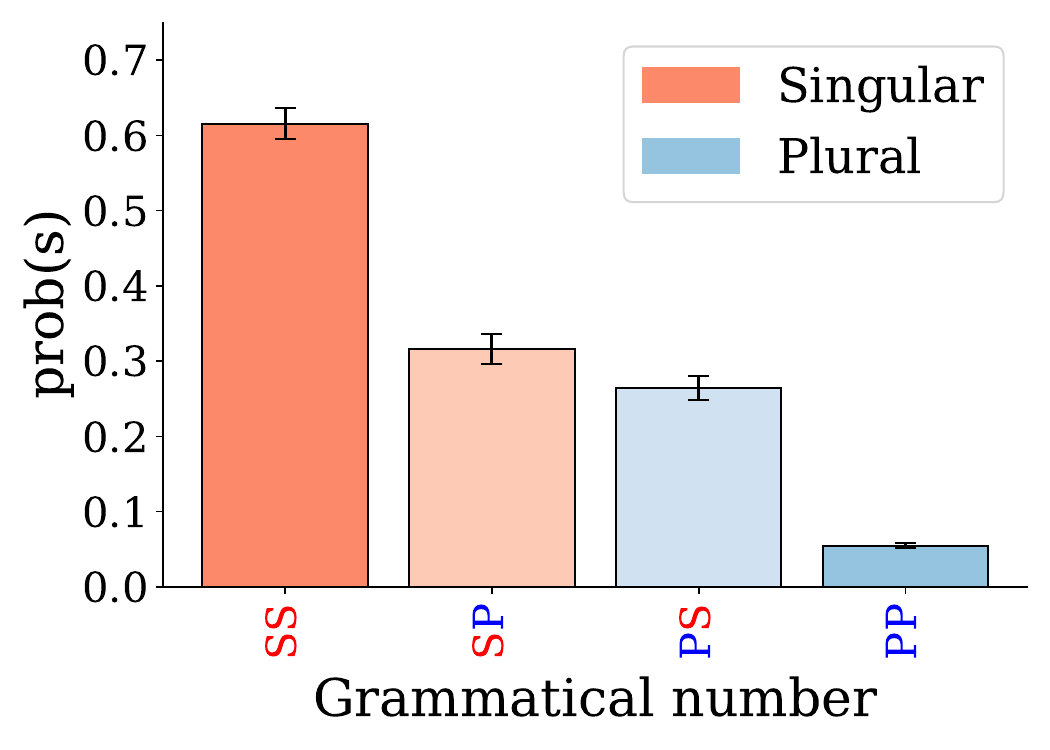}
  \caption{Probability for predicting the token `s' for different grammatical number combinations of the subject and the attractor. }
  \label{fig:cong_pred}
\end{figure}

\begin{figure}[H]
  \centering
  \includegraphics[width=0.5\columnwidth]{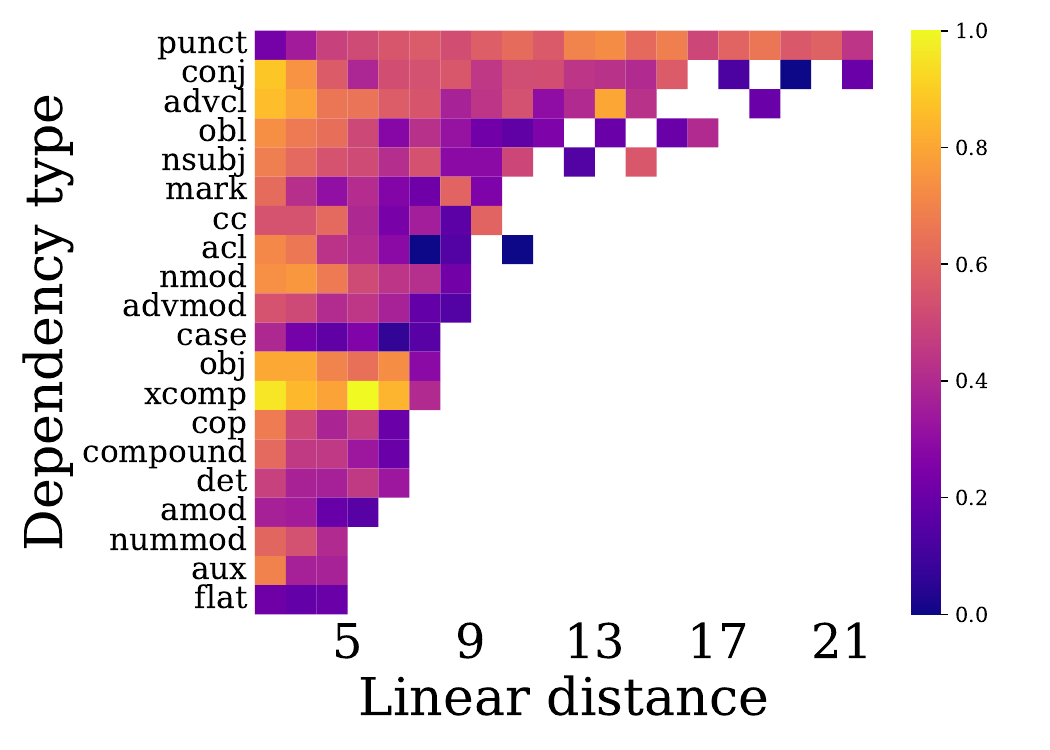}
  \caption{Polar probe accuracy for dependencies by their linear distance and dependency type.}
  \label{fig:matrix}
\end{figure}

\begin{figure}[H]
  \centering
  \includegraphics[width=0.5\columnwidth]{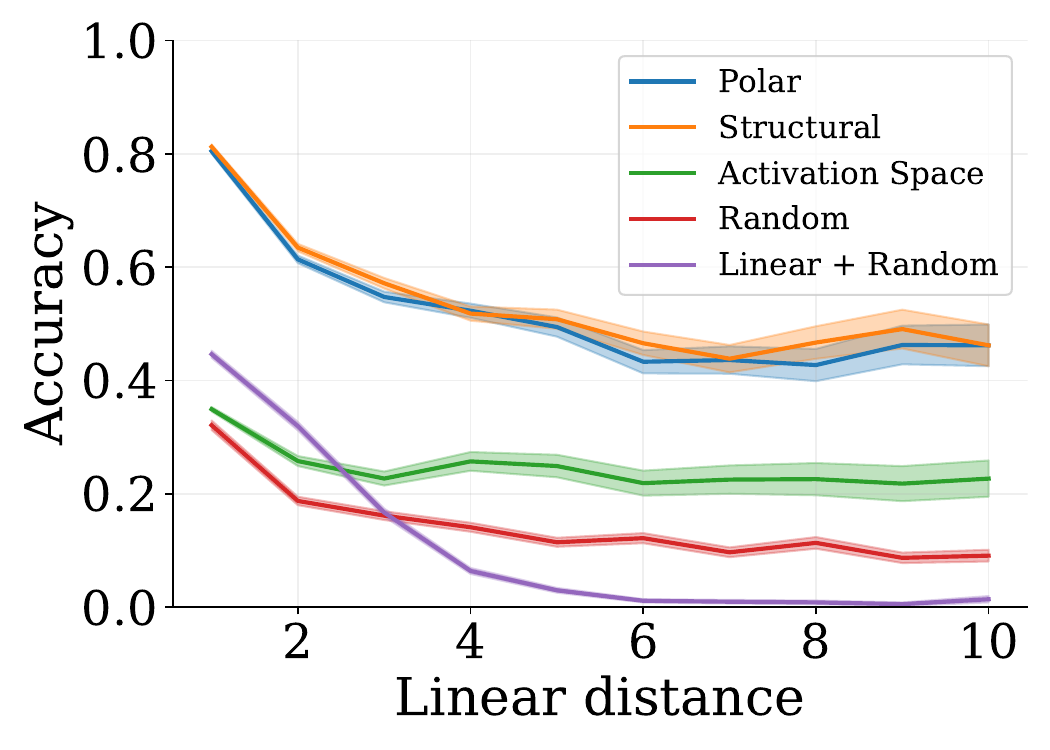}
  \caption{Accuracy as a function of linear distance in UD-EWT}
  \label{lengths_with_control}
\end{figure}

\begin{figure}[H]
  \centering
  \includegraphics[width=0.5\columnwidth]{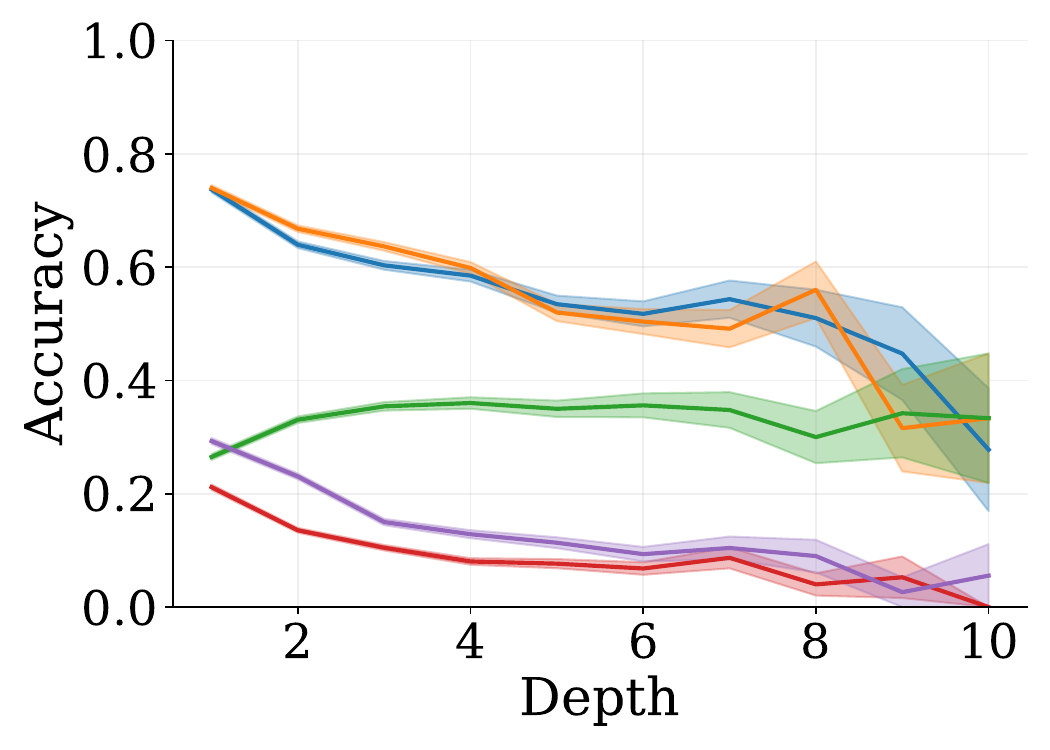}
  \caption{Accuracy as a function of syntactic depth in UD-EWT}
  \label{depths_with_control}
\end{figure}

\begin{figure}[H]
  \centering
  \includegraphics[width=0.5\columnwidth]{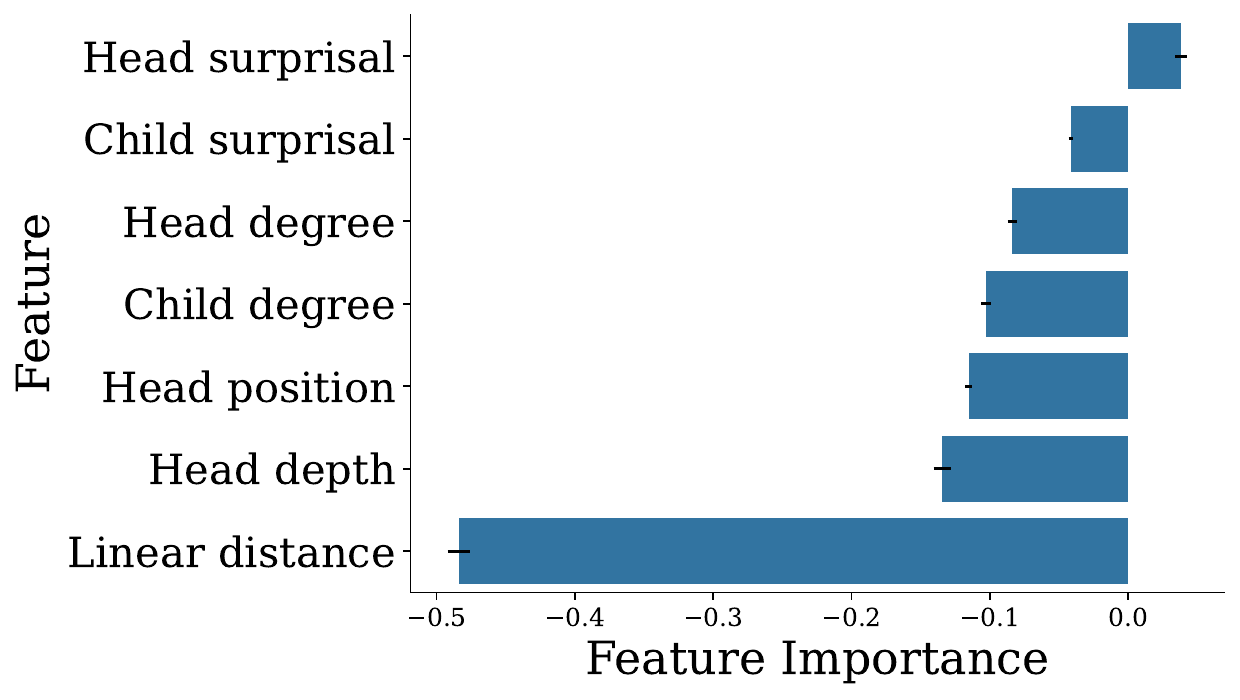}
  \caption{Feature importance values in Random Forest model with additional features.}
  \label{more_features}
\end{figure}

\break
\section{Llama2-7B}


\begin{figure*}[ht]
    \centering
    \begin{subfigure}[b]{0.32\textwidth}
        \centering
        \includegraphics[width=\textwidth]{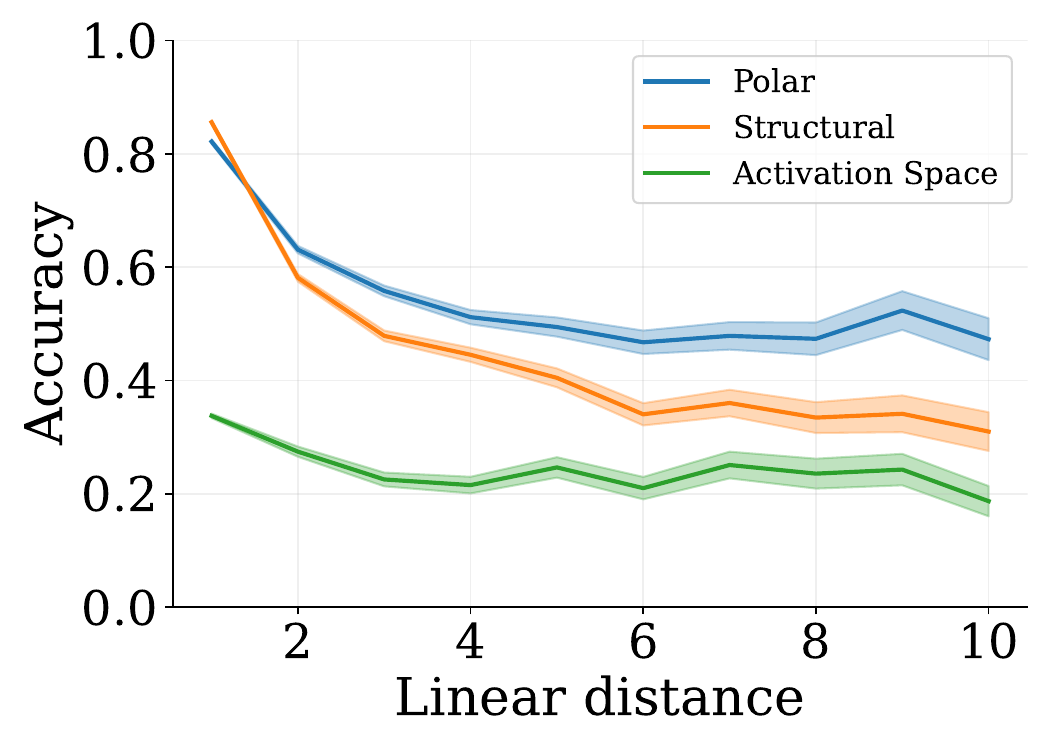}
        \phantomcaption
        \label{fig:lengths_llama}
    \end{subfigure}
    \hfill
    \begin{subfigure}[b]{0.32\textwidth}
        \centering
        \includegraphics[width=\textwidth]{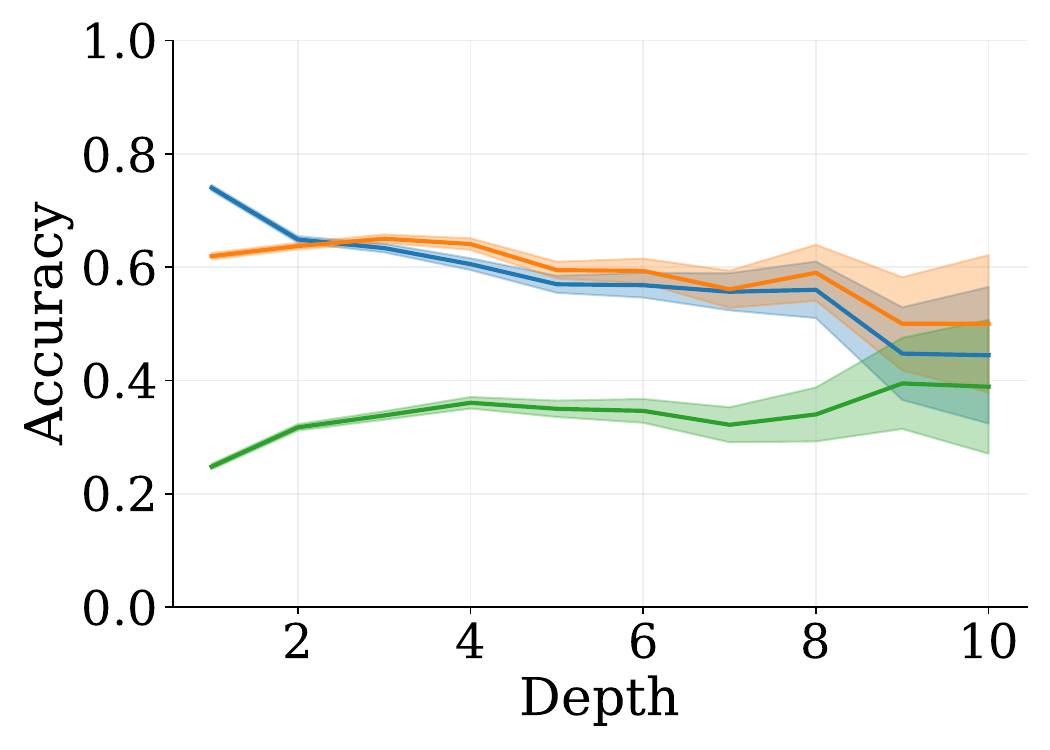}
        \phantomcaption
        \label{fig:depths_llama}
    \end{subfigure}
    \hfill
    \begin{subfigure}[b]{0.32\textwidth}
        \centering
        \includegraphics[width=\textwidth]{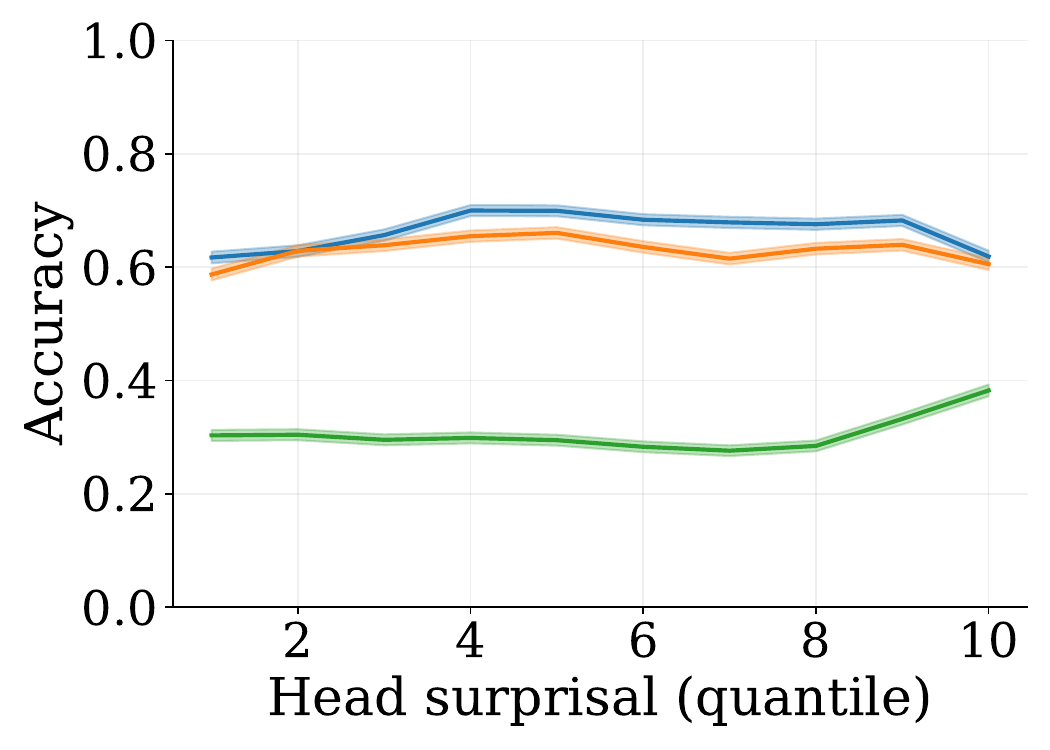}
        \phantomcaption
        \label{fig:surprisal_llama}
    \end{subfigure}
    
    \vspace{-0.4cm} 
    
    \begin{subfigure}[b]{0.32\textwidth}
        \centering
        \includegraphics[width=\textwidth]{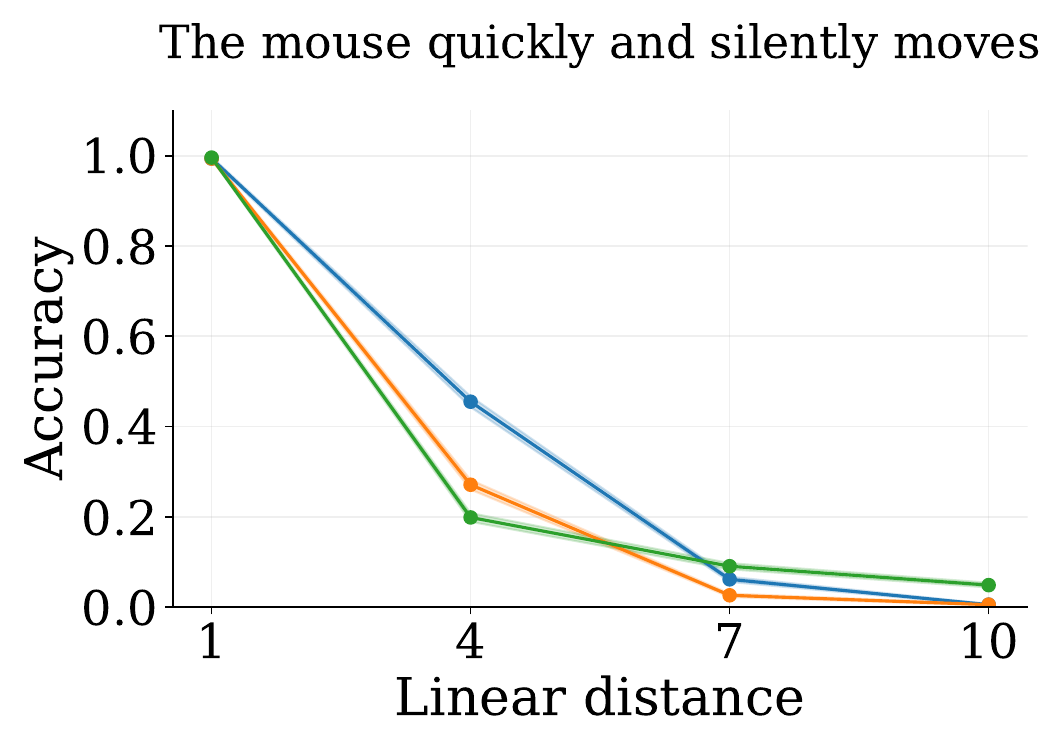}
        \phantomcaption
        \label{fig:lengths_controlled_llama}
    \end{subfigure}
    \hfill
    \begin{subfigure}[b]{0.32\textwidth}
        \centering
        \includegraphics[width=\textwidth]{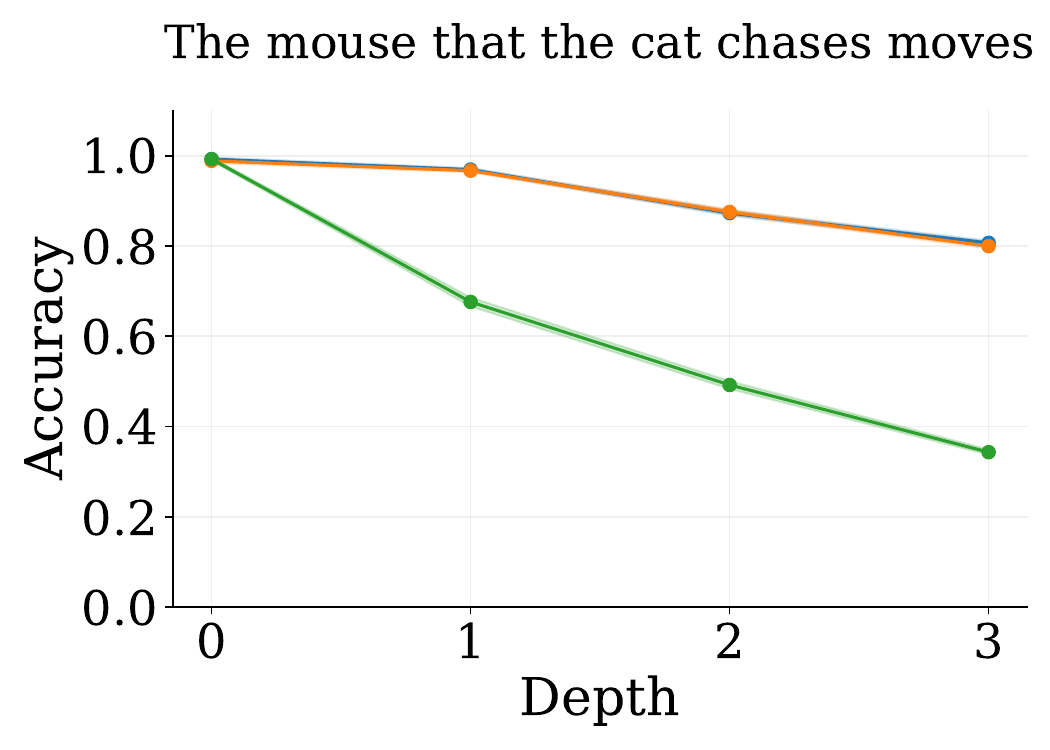}
        \phantomcaption
        \label{fig:depths_controlled_llama}
    \end{subfigure}
    \hfill
    \begin{subfigure}[b]{0.32\textwidth}
        \centering
        \includegraphics[width=\textwidth]{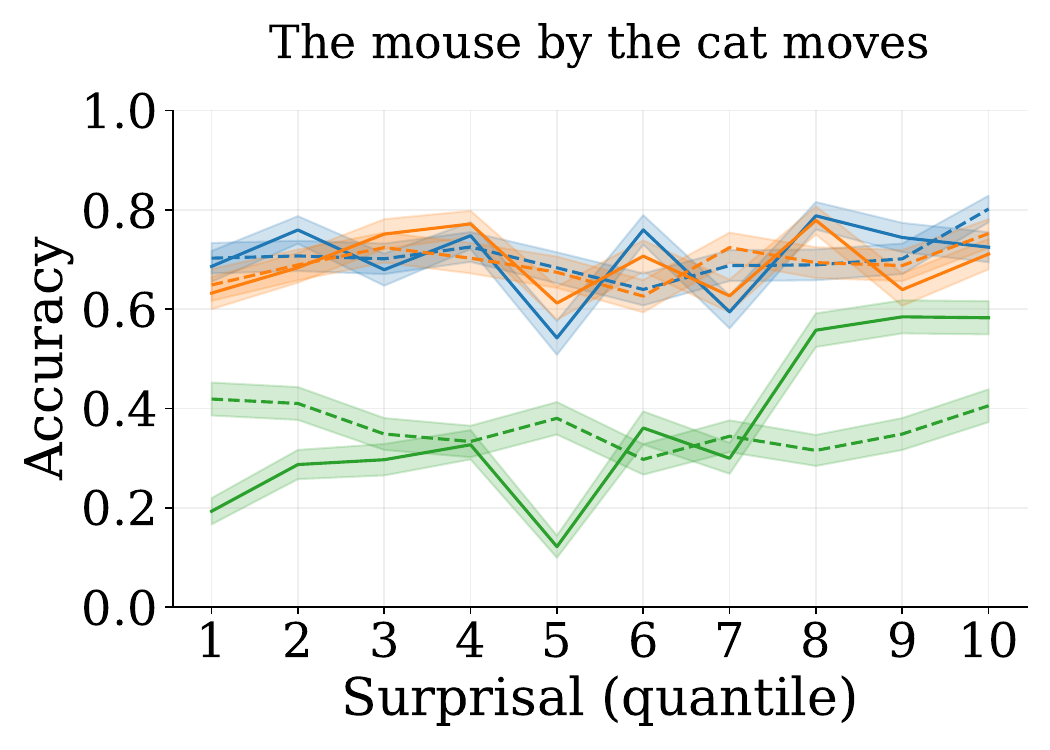}
        \phantomcaption
        \label{fig:surprisal_controlled_llama}
    \end{subfigure}
    \vspace{-0.5cm}
    
    \label{fig:marginal_acc_llama}
    \caption{}
\end{figure*}


\begin{figure*}[ht]
    \centering
    \begin{subfigure}[b]{0.49\textwidth}
        \centering
        \includegraphics[width=\textwidth]{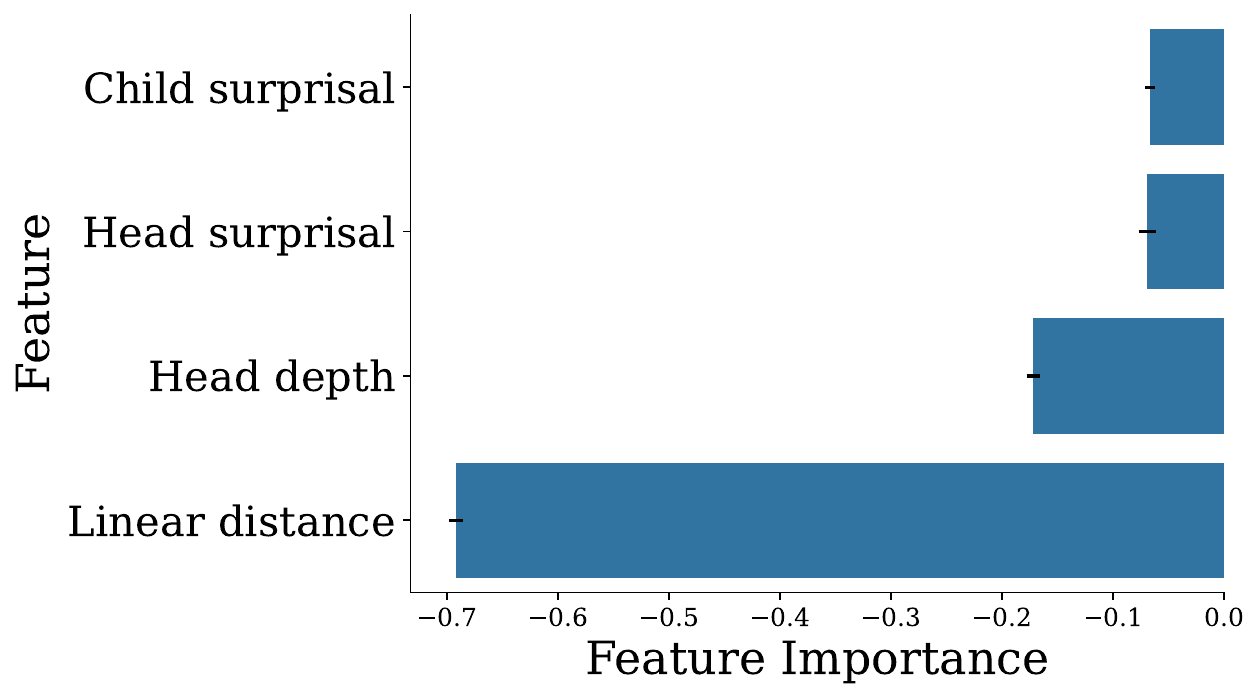}
        \phantomcaption
        \label{fig:coeffs_llama_less}
    \end{subfigure}
    \hfill
    \begin{subfigure}[b]{0.49\textwidth}
        \centering
        \includegraphics[width=\textwidth]{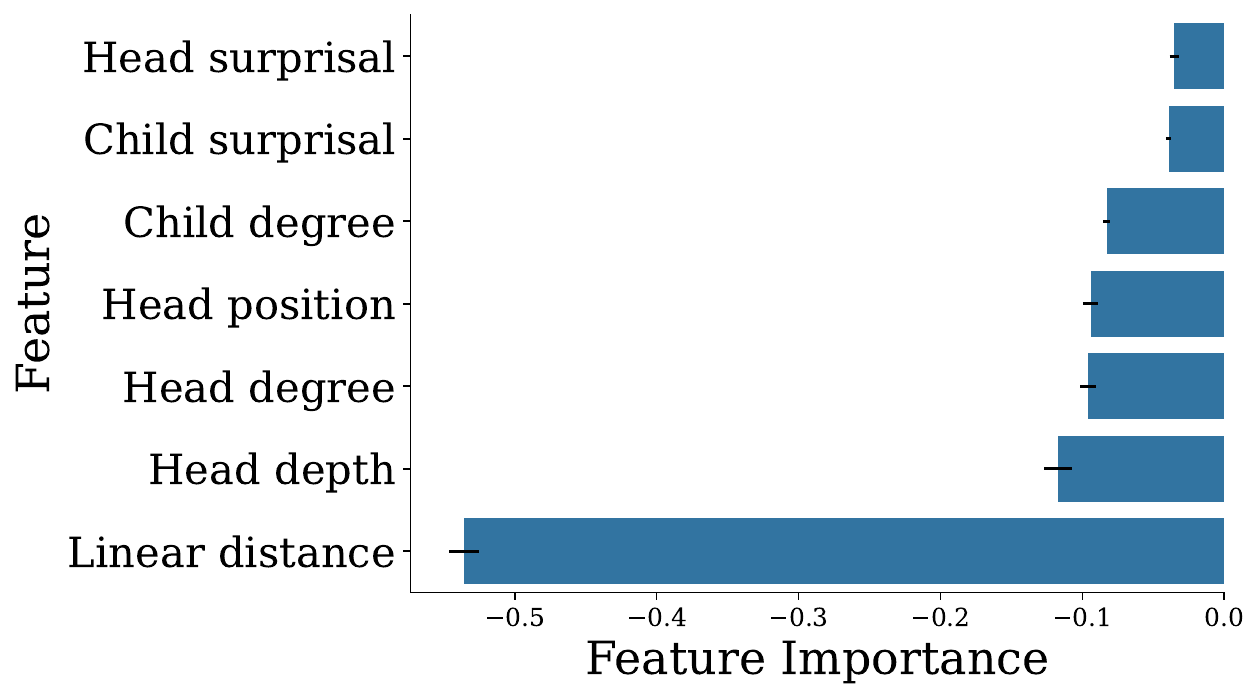}
        \phantomcaption
        \label{fig:coeffs_llama_more}
    \end{subfigure}
    \caption{}
    \label{fig:coeffs_llama}
\end{figure*}

\break
\begin{figure*}[ht]
  \centering
  \begin{subfigure}[b]{0.48\textwidth}
    \includegraphics[width=\textwidth]{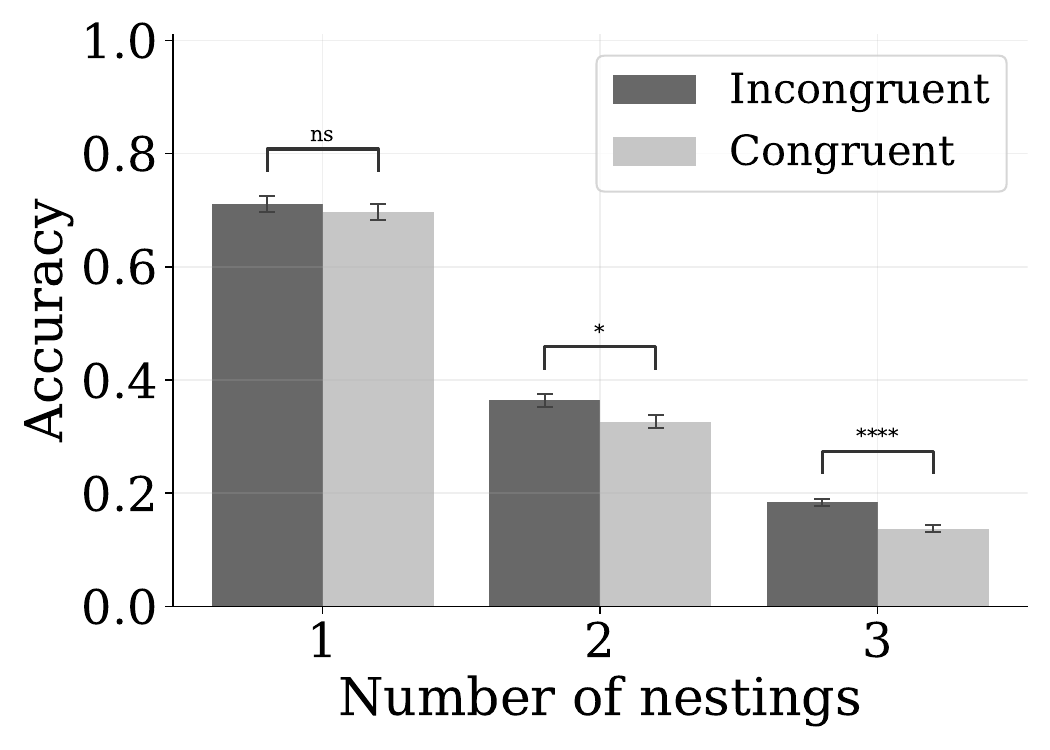}
    \phantomcaption
    \label{fig:cong_linear_llama}
  \end{subfigure}
  \hfill
  \begin{subfigure}[b]{0.48\textwidth}
    \includegraphics[width=\textwidth]{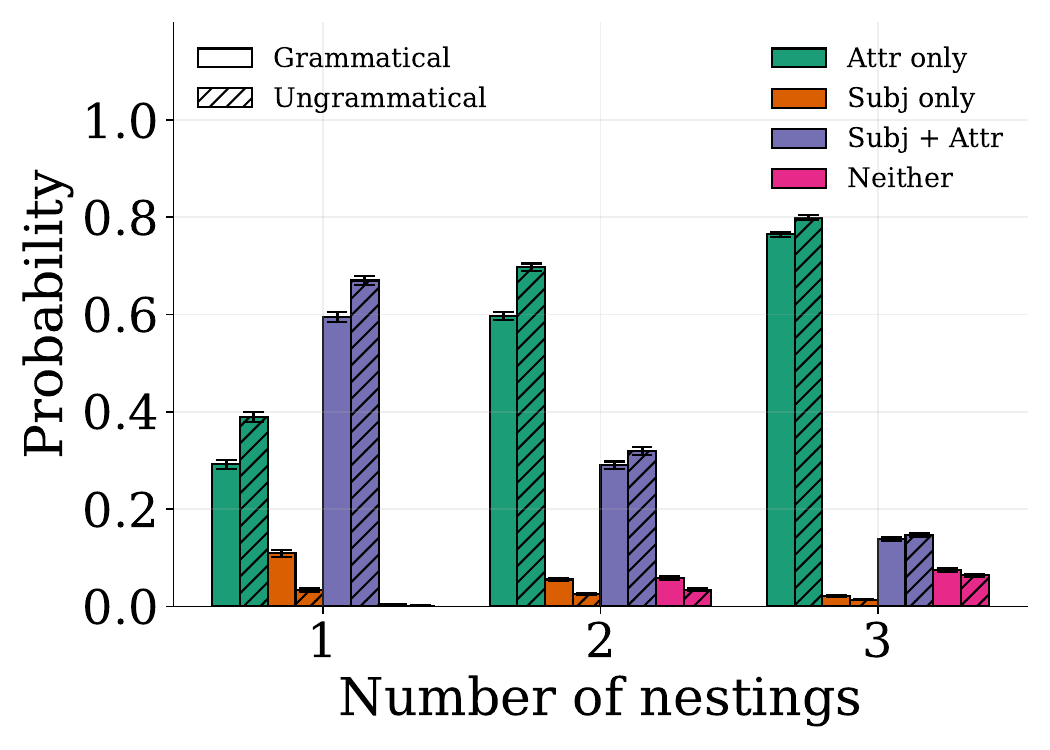}
    \phantomcaption
    \label{fig:flips_llama}
  \end{subfigure}
  
  \vspace{-0.5cm} 
  
  \begin{subfigure}[b]{0.48\textwidth}
    \includegraphics[width=\textwidth]{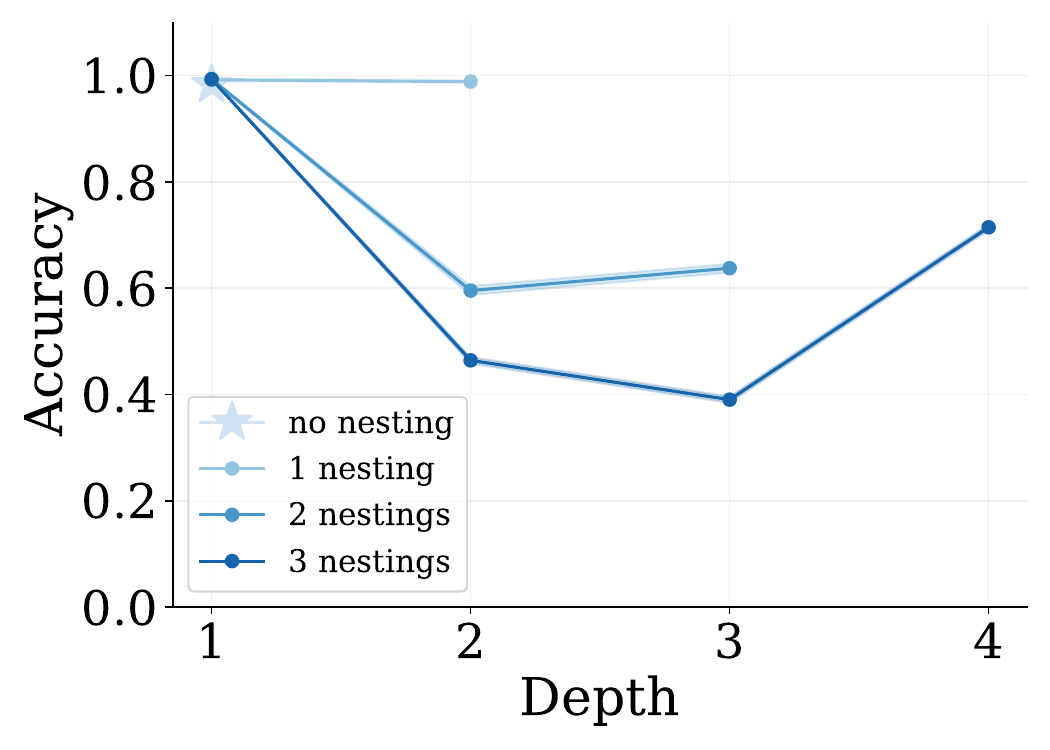}
    \phantomcaption
    \label{fig:controlled_depths_rb_llama}
  \end{subfigure}
  \hfill
  \begin{subfigure}[b]{0.48\textwidth}
    \includegraphics[width=\textwidth]{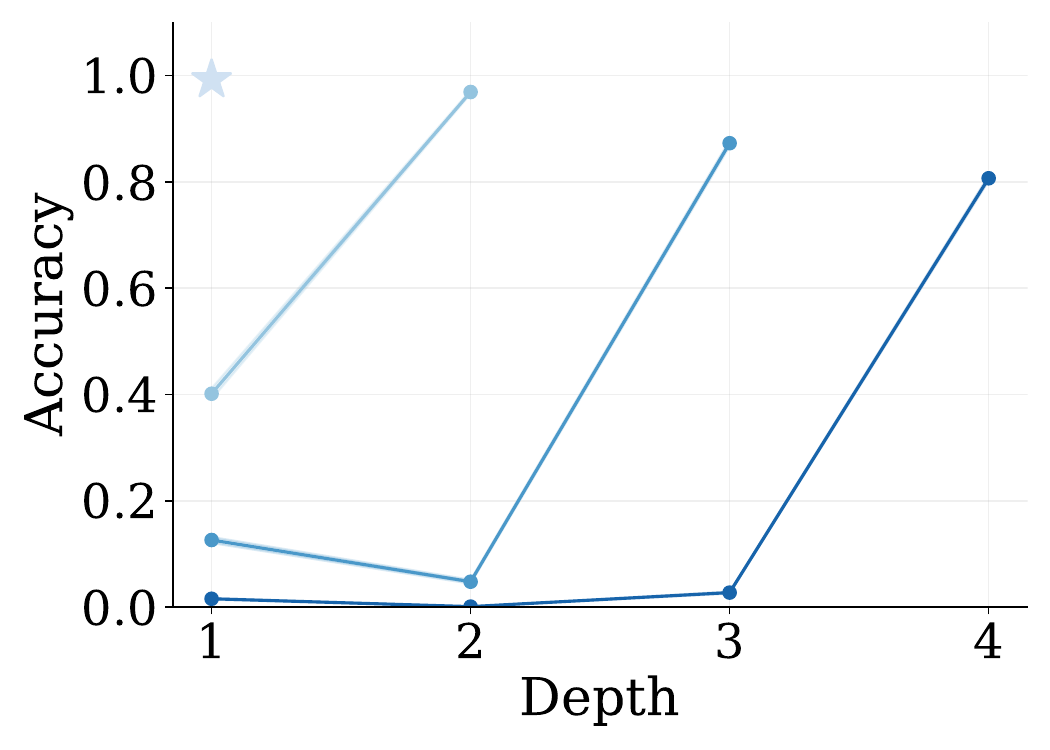}
    \phantomcaption
    \label{fig:controlled_depths_ce_llama}
  \end{subfigure}
  
  \vspace{-0.5cm}
  \label{fig:2x2_figure_llama}
  \caption{}

\end{figure*}

\newpage
\section{Mistral-7B}
\subsection{Structural Probe}

\begin{figure*}[ht]
  \centering
  \begin{subfigure}[b]{0.48\textwidth}
    \includegraphics[width=\textwidth]{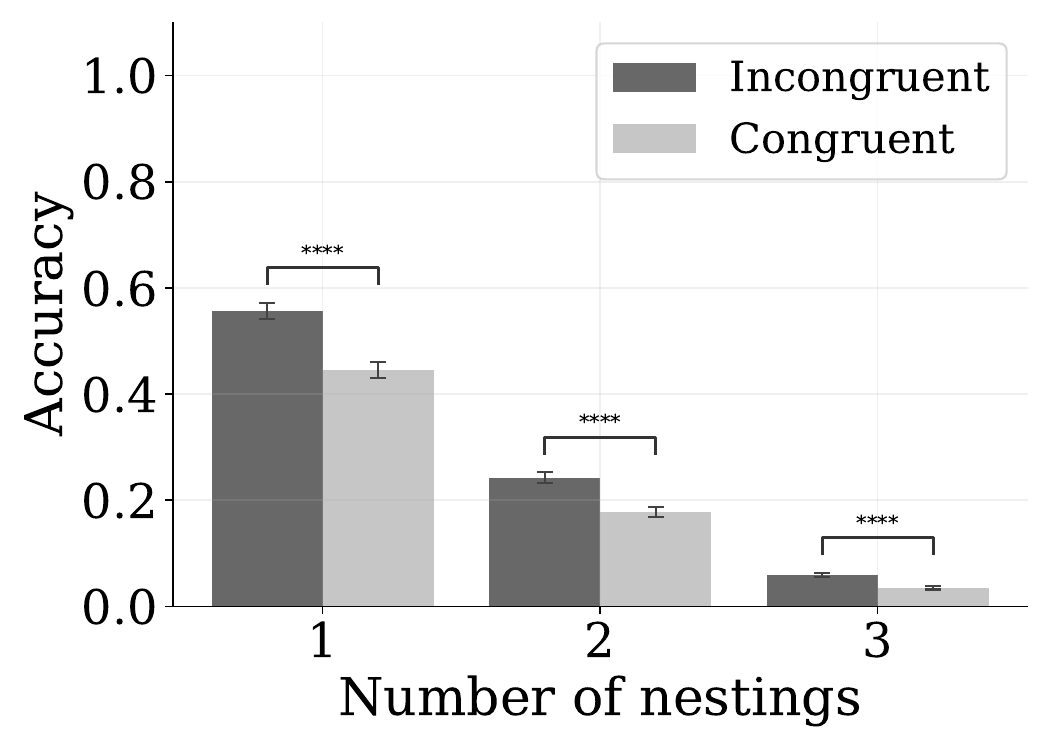}
    \phantomcaption
    \label{fig:cong_linear_mistral_structural}
  \end{subfigure}
  \hfill
  \begin{subfigure}[b]{0.48\textwidth}
    \includegraphics[width=\textwidth]{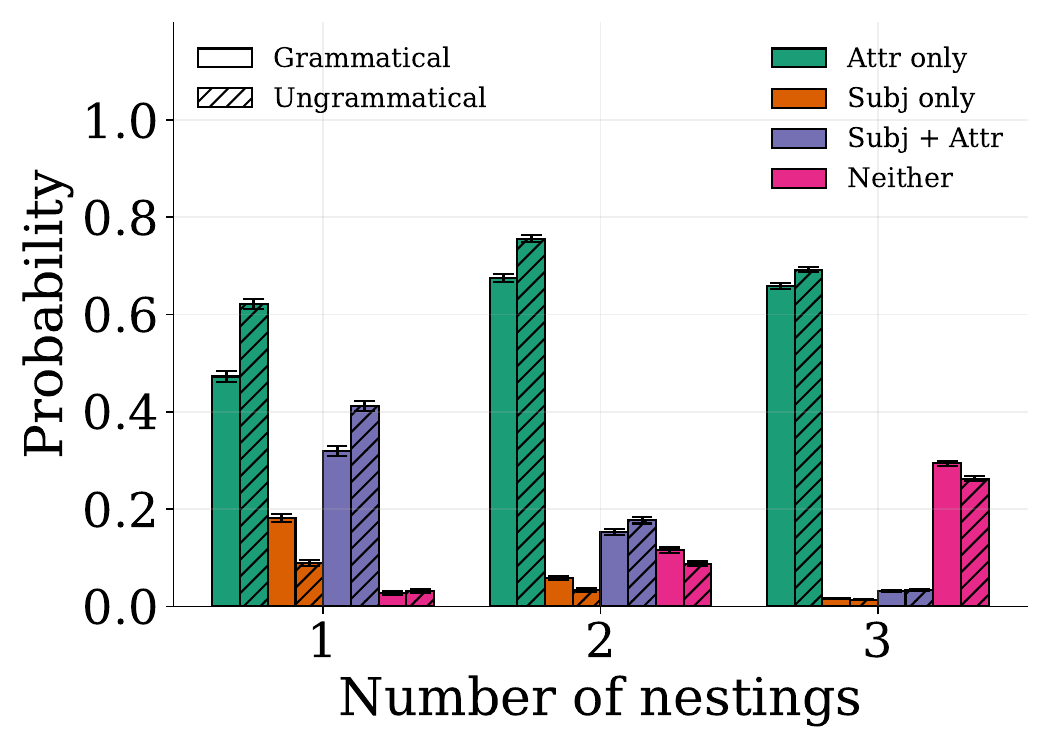}
    \phantomcaption
    \label{fig:flips_mistral_structural}
  \end{subfigure}
  
  \vspace{-0.5cm} 
  
  \begin{subfigure}[b]{0.48\textwidth}
    \includegraphics[width=\textwidth]{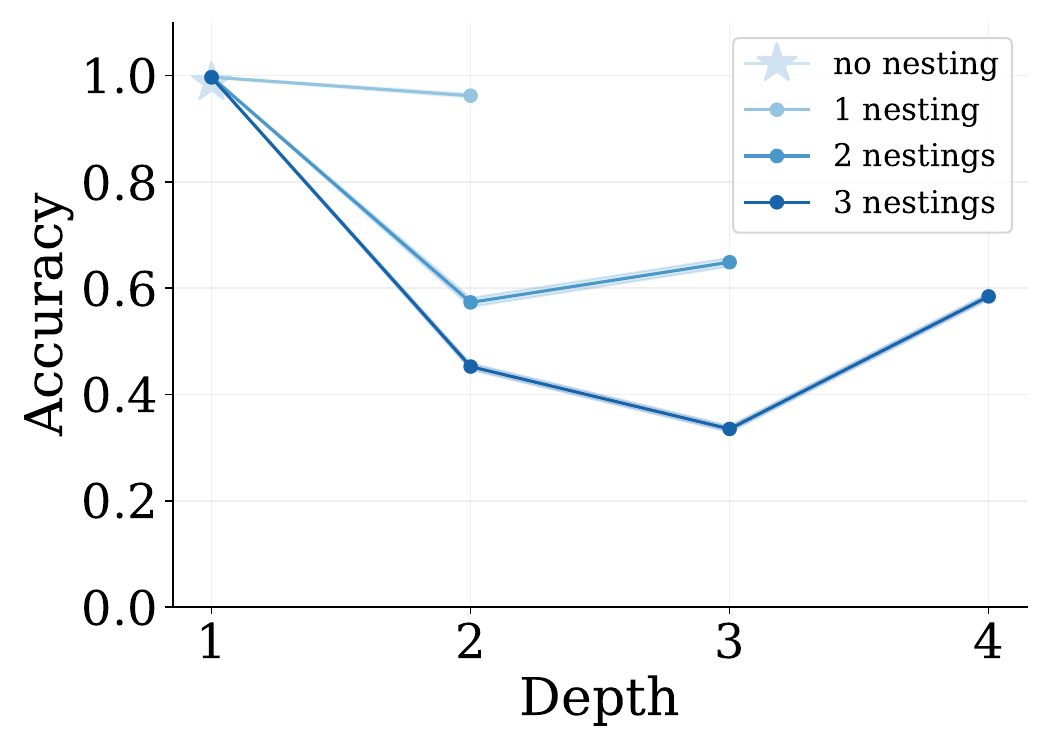}
    \phantomcaption
    \label{fig:controlled_depths_rb_mistral_structural}
  \end{subfigure}
  \hfill
  \begin{subfigure}[b]{0.48\textwidth}
    \includegraphics[width=\textwidth]{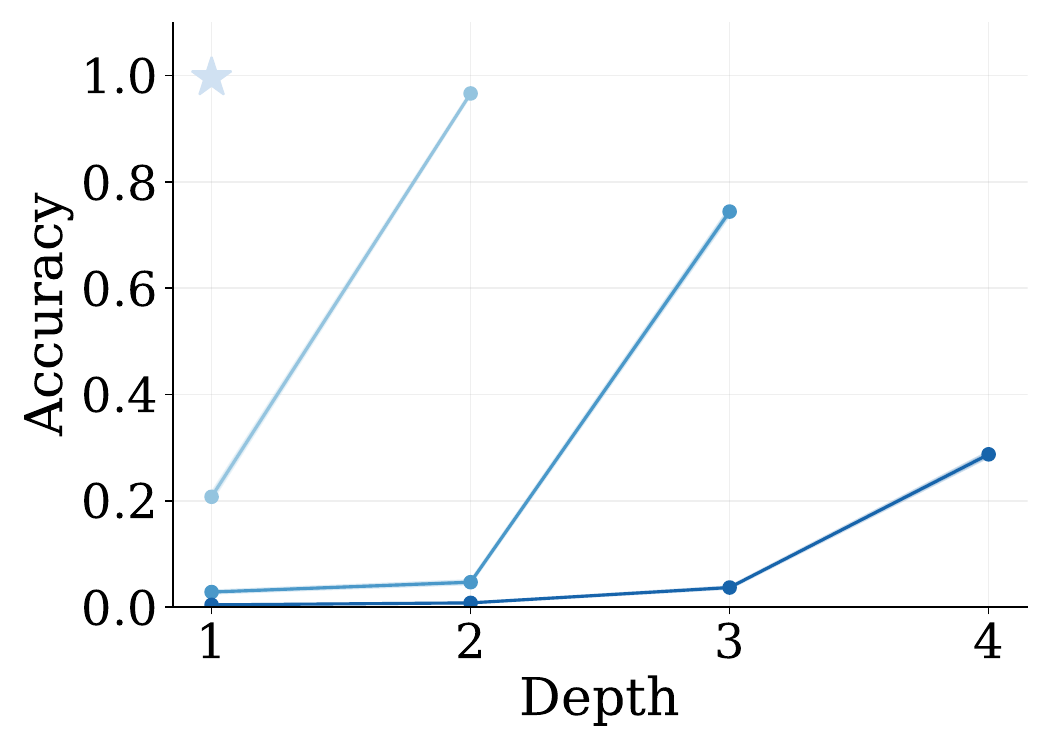}
    \phantomcaption
    \label{fig:controlled_depths_ce_mistral_structural}
  \end{subfigure}
  
  \vspace{-0.5cm}
  \label{fig:2x2_figure_mistal_structural}
  \caption{}

\end{figure*}

\newpage
\section{BERT-large}


\begin{figure*}[ht]
    \centering
    \begin{subfigure}[b]{0.49\textwidth}
        \centering
        \includegraphics[width=\textwidth]{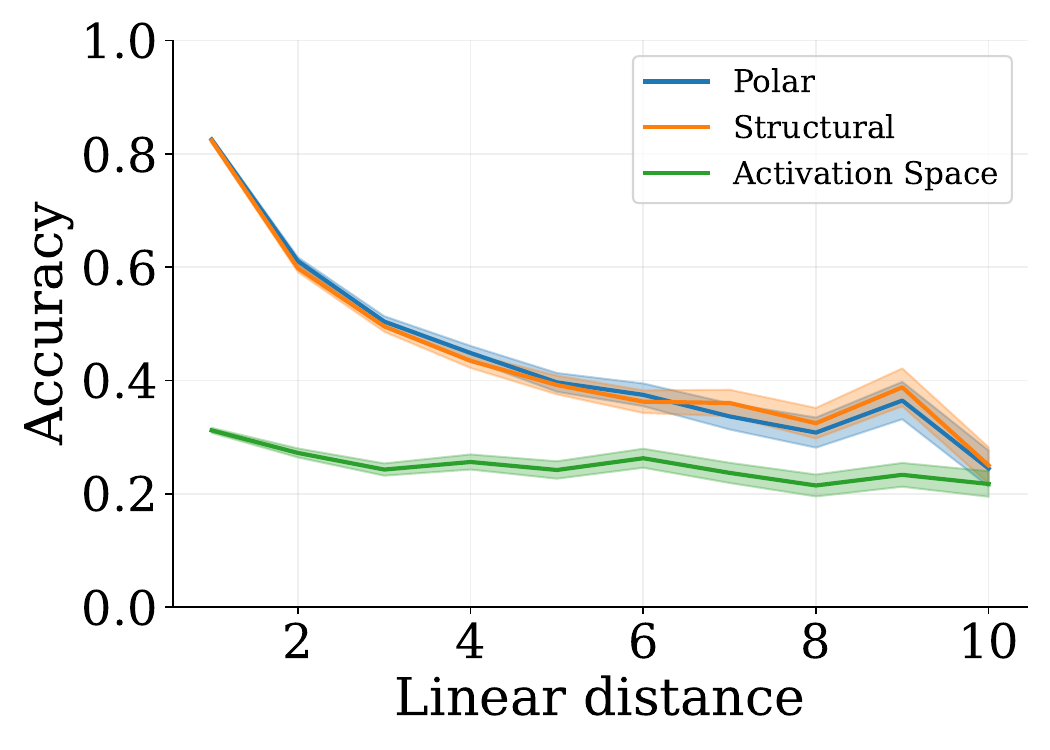}
        \phantomcaption
        \label{fig:lengths_bert-large}
    \end{subfigure}
    \hfill
    \begin{subfigure}[b]{0.49\textwidth}
        \centering
        \includegraphics[width=\textwidth]{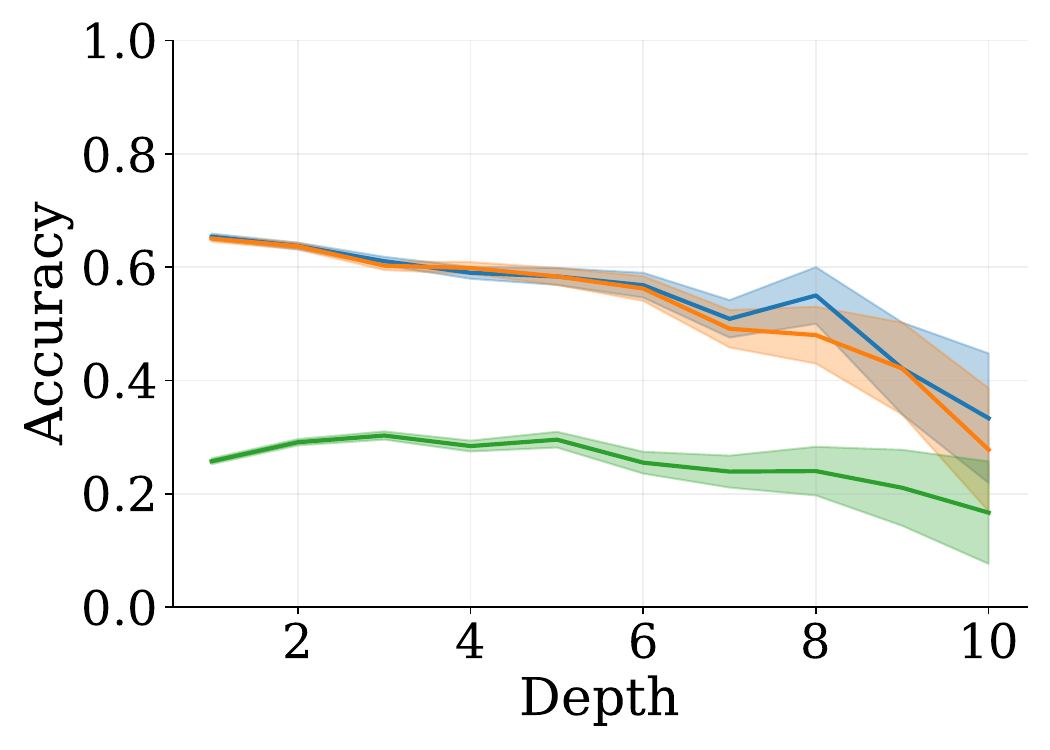}
        \phantomcaption
        \label{fig:depths_bert-large}
    \end{subfigure}
    
    \vspace{-0.4cm} 
    
    \begin{subfigure}[b]{0.49\textwidth}
        \centering
        \includegraphics[width=\textwidth]{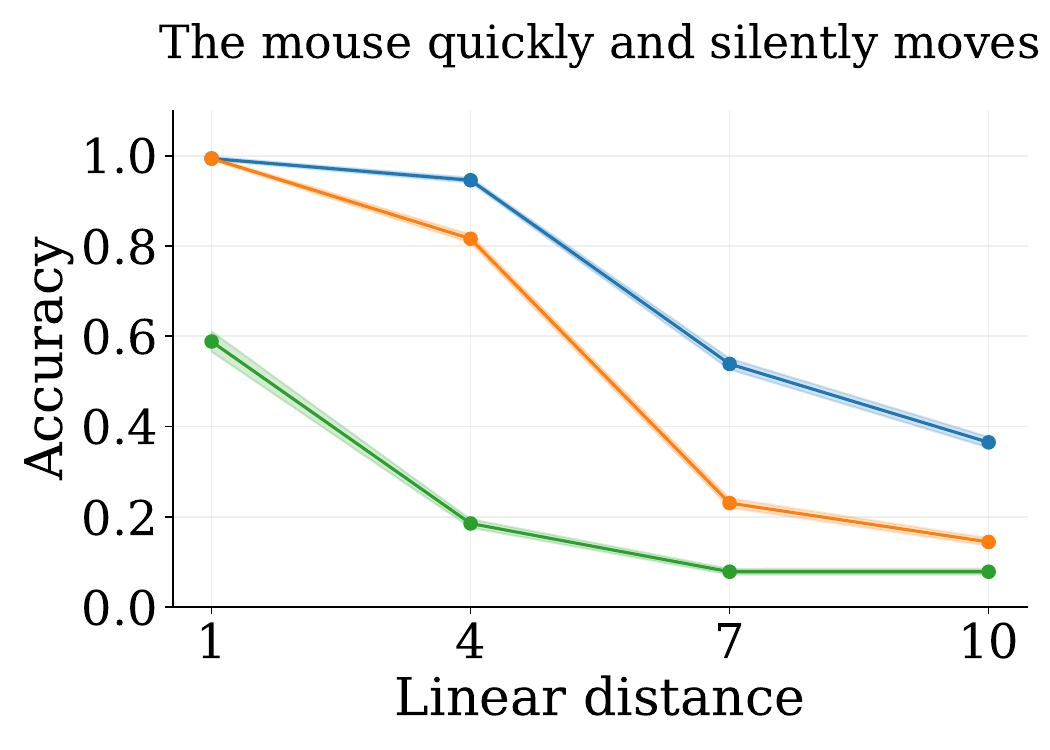}
        \phantomcaption
        \label{fig:lengths_controlled_bert-large}
    \end{subfigure}
    \hfill
    \begin{subfigure}[b]{0.49\textwidth}
        \centering
        \includegraphics[width=\textwidth]{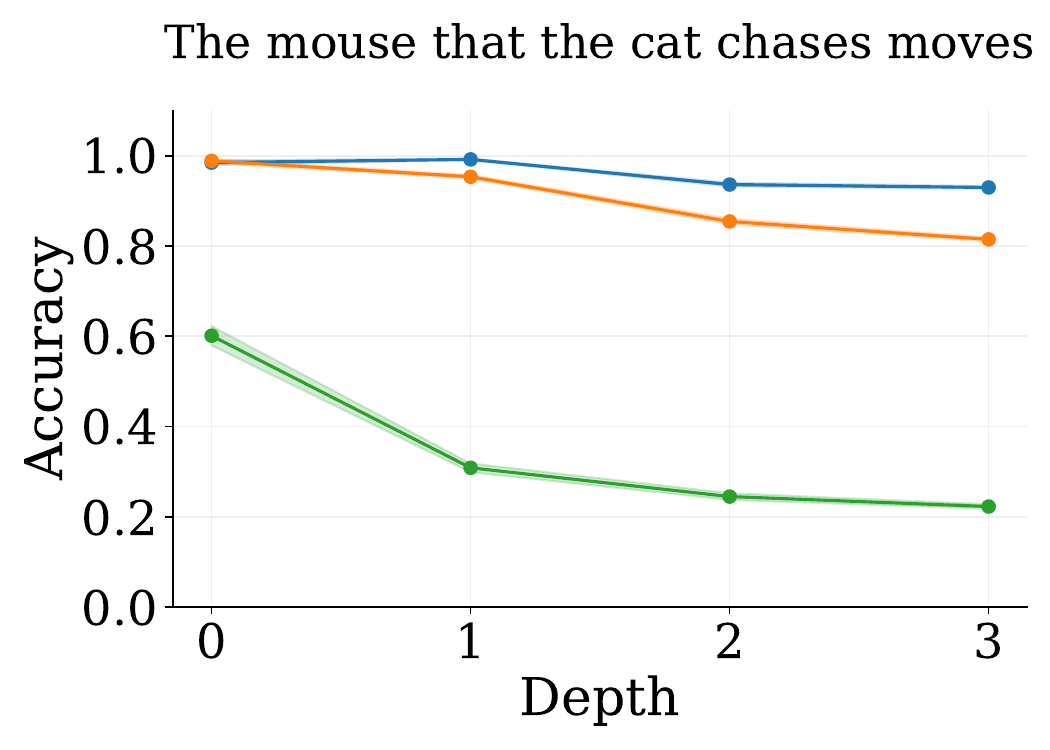}
        \phantomcaption
        \label{fig:depths_controlled_bert-large}
    \end{subfigure}
    \vspace{-0.5cm}
    
    \label{fig:marginal_acc_bert-large}
    \caption{}

\end{figure*}

\begin{figure*}[ht]
  \centering
  \begin{subfigure}[b]{0.48\textwidth}
    \includegraphics[width=\textwidth]{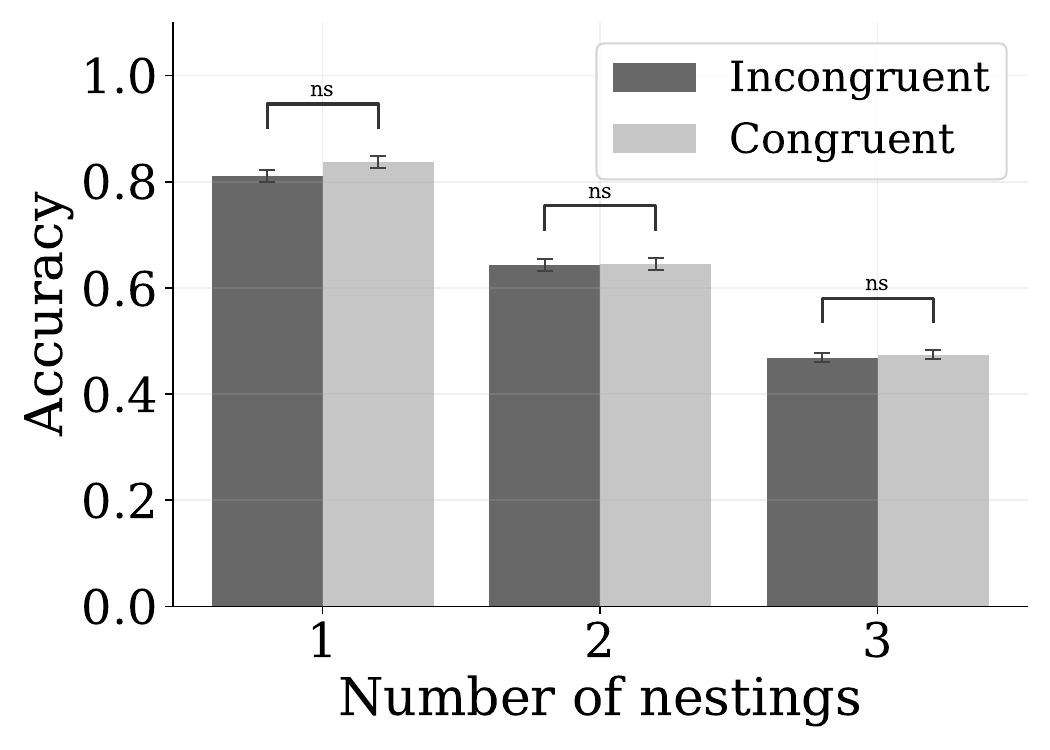}
    \phantomcaption
    \label{fig:cong_linear_bert-large}
  \end{subfigure}
  \hfill
  \begin{subfigure}[b]{0.48\textwidth}
    \includegraphics[width=\textwidth]{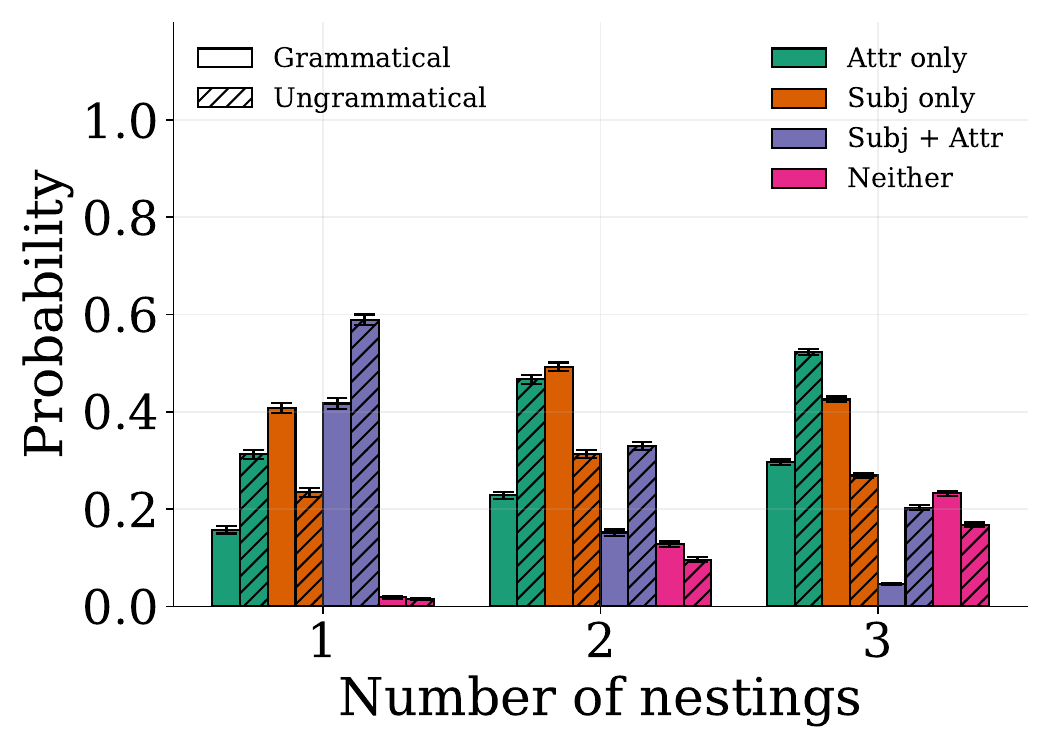}
    \phantomcaption
    \label{fig:flips_bert-large}
  \end{subfigure}
  
  \vspace{-0.5cm} 
  
  \begin{subfigure}[b]{0.48\textwidth}
    \includegraphics[width=\textwidth]{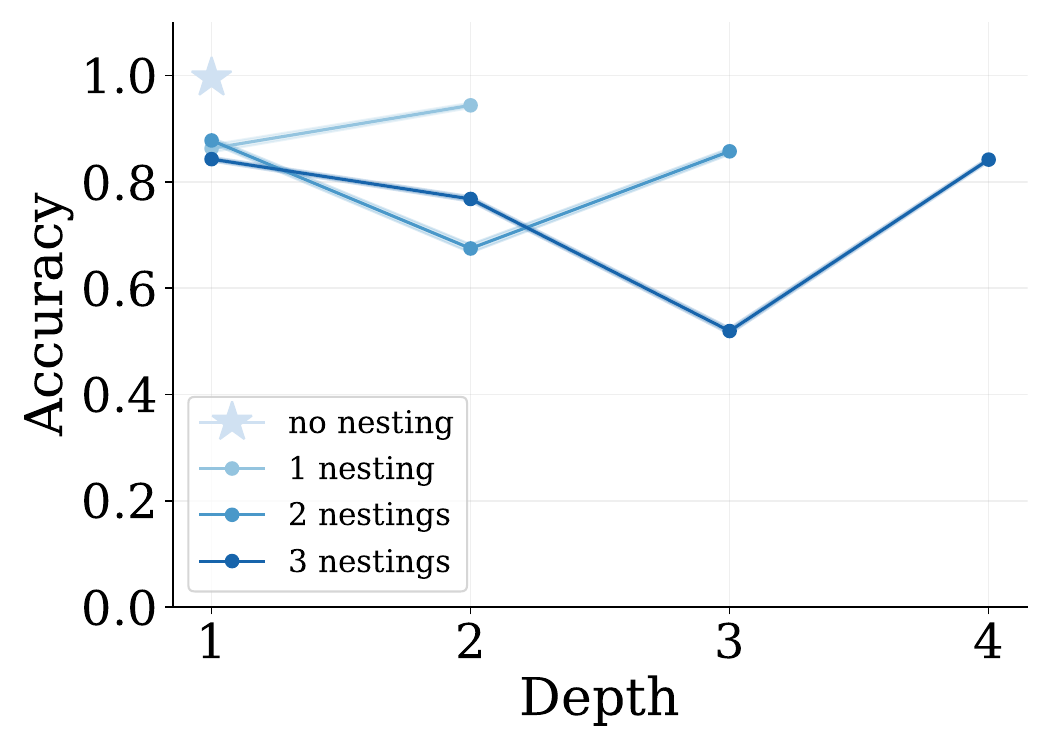}
    \phantomcaption
    \label{fig:controlled_depths_rb_bert-large}
  \end{subfigure}
  \hfill
  \begin{subfigure}[b]{0.48\textwidth}
    \includegraphics[width=\textwidth]{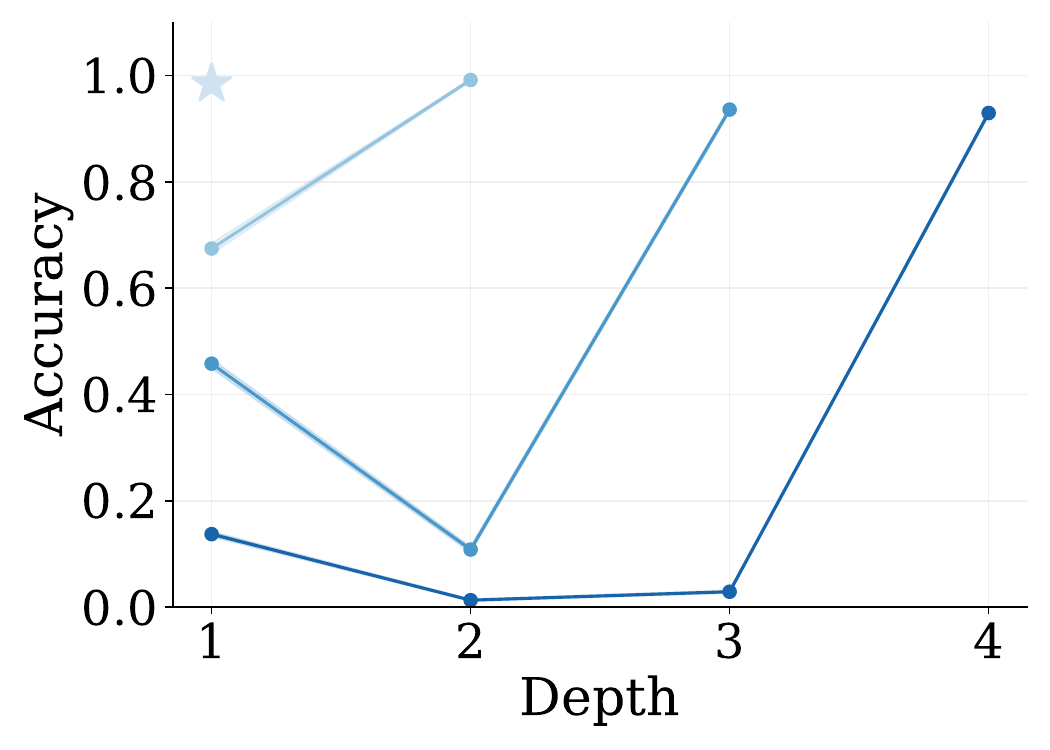}
    \phantomcaption
    \label{fig:controlled_depths_ce_bert-large}
  \end{subfigure}
  
  \vspace{-0.5cm}
  \label{fig:2x2_figure_bert-large}
  \caption{}

\end{figure*}

\newpage

\section{Controlled Dataset}

\subsection{Prepositional Phrase}

\begin{table}[H]
\centering
\begin{tabular}{|
  >{\tiny}p{8cm}|
  >{\tiny}p{1cm}|
  >{\tiny}p{1cm}|
  >{\tiny}p{1cm}|
}

\hline
\textbf{Sentence} & \textbf{Structure} & \textbf{Attractors} & \textbf{Fillers} \\
\hline
some neighbours by some boys rest & pp & 1 & 0 \\
the teacher below some friends around some bakers plays & pp & 2 & 0 \\
a tailor above the boy near a teacher by the kids rests & pp & 3 & 0 \\
the girls past a boy quickly and carefully stand & pp & 1 & 1 \\
the grandma below a friend gently eagerly playfully happily and nervously plays & pp & 1 & 2 \\
a kid behind a soldier near the waiters playfully and silently sits & pp & 2 & 1 \\
\hline
\end{tabular}
\caption{Examples of PP sentences in the controlled dataset.}
\label{tab:PP_examples}
\end{table}

\subsection{Center Embedding}

\begin{table}[H]
\centering
\begin{tabular}{|
  >{\tiny}p{8cm}|
  >{\tiny}p{1cm}|
  >{\tiny}p{1cm}|
  >{\tiny}p{1cm}|
}

\hline
\textbf{Sentence} & \textbf{Structure} & \textbf{Attractors} & \textbf{Fillers} \\
\hline
some teachers that the tailor admires walk & ce & 1 & 0 \\
some players that the neighbors that a mom calls find rest & ce & 2 & 0 \\
the teacher that some moms that a tailor that the guests show finds lift stands & ce & 3 & 0 \\
the grandmas that some engineers loudly carefully and boldly help smile & ce & 1 & 1 \\
an artist that the player swiftly happily joyfully silently gently quietly and nervously hears sits & ce & 1 & 2 \\
the artist that the neighbor that the soldier joyfully boldly and quickly kicks meets walks & ce & 2 & 1 \\
\hline
\end{tabular}
\caption{Examples of CE sentences in the controlled dataset.}
\label{tab:CE_examples}
\end{table}

\subsection{Right Branching}

\begin{table}[H]
\centering
\begin{tabular}{|
  >{\tiny}p{8cm}|
  >{\tiny}p{1cm}|
  >{\tiny}p{1cm}|
  >{\tiny}p{1cm}|
}

\hline
\textbf{Sentence} & \textbf{Structure} & \textbf{Attractors} & \textbf{Fillers} \\
\hline
a friend assumes that the baker predicts & rb & 1 & 0 \\
a singer imagines that the guests feel that the neighbor realizes & rb & 2 & 0 \\
the girls realize that a writer predicts that some teachers think that an engineer imagines & rb & 3 & 0 \\
a boy feels that the sad shy and brave engineer expects & rb & 1 & 1 \\
a baker notices that the friendly young old angry funny lazy and wild dads believe & rb & 1 & 2 \\
some grandmas feel that some neighbors notice that the funny kind and shy boys know & rb & 2 & 1 \\
\hline
\end{tabular}
\caption{Examples of RB sentences in the controlled dataset.}
\label{tab:RB_examples}
\end{table}

\end{document}